\documentclass[sigconf]{acmart}
\copyrightyear{2021}
\acmYear{2021} 
\setcopyright{iw3c2w3}
\acmConference[WWW '21]{Proceedings of the Web Conference 2021}{April 19--23, 2021}{Ljubljana, Slovenia} 
\acmBooktitle{Proceedings of the Web Conference 2021 (WWW '21), April 19--23, 2021, Ljubljana, Slovenia}
\acmPrice{}
\acmDOI{10.1145/3442381.3449988}
\acmISBN{978-1-4503-8312-7/21/04}
\usepackage{stfloats}
\usepackage{subfigure}
\usepackage{tabularx}
\usepackage{array}
\usepackage{booktabs}
\usepackage{setspace}
\usepackage{url}
\usepackage{enumerate}

\AtBeginDocument{%
  \providecommand\BibTeX{{%
    \normalfont B\kern-0.5em{\scshape i\kern-0.25em b}\kern-0.8em\TeX}}}

\begin{document}

\title{Using Prior Knowledge to Guide BERT's Attention in Semantic Textual Matching Tasks}

\author{Tingyu Xia}
\affiliation{School of Artificial Intelligence, Jilin University\\ 
Key Laboratory of Symbolic Computation and Knowledge Engineering, Jilin University}
\email{xiaty19@mails.jlu.edu.cn} 

\author{Yue Wang}
\affiliation{School of Information and Library Science\\University of North Carolina at Chapel Hill}
\email{wangyue@unc.edu} 

\author{Yuan Tian}
\affiliation{School of Artificial Intelligence, Jilin University\\ 
Key Laboratory of Symbolic Computation and Knowledge Engineering, Jilin University}
\authornote{Joint Corresponding Authors}
\email{yuantian@jlu.edu.cn} 

\author{Yi Chang}
\affiliation{School of Artificial Intelligence, Jilin University\\ 
International Center of Future Science, Jilin University\\
Key Laboratory of Symbolic Computation and Knowledge Engineering, Jilin University}
\authornotemark[1]
\email{yichang@jlu.edu.cn}

\renewcommand{\shortauthors}{Xia and Wang, et al.}

\begin{abstract}
We study the problem of incorporating prior knowledge into a deep Transformer-based model, i.e., Bidirectional Encoder Representations from Transformers (BERT), to enhance its performance on semantic textual matching tasks. By probing and analyzing what BERT has already known when solving this task, we obtain better understanding of what task-specific knowledge BERT needs the most and where it is most needed. The analysis further motivates us to take a different approach than most existing works. Instead of using prior knowledge to create a new training task for fine-tuning BERT, we directly inject knowledge into BERT's multi-head attention mechanism. This leads us to a simple yet effective approach that enjoys fast training stage as it saves the model from training on additional data or tasks other than the main task. Extensive experiments demonstrate that the proposed knowledge-enhanced BERT is able to consistently improve semantic textual matching performance over the original BERT model, and the performance benefit is most salient when training data is scarce.
\end{abstract}


\maketitle
\section{Introduction}
Measuring semantic similarity between two pieces of text  is a fundamental and important task in natural language understanding. Early works on this task often leverage knowledge resources such as WordNet \cite{miller1995wordnet} and UMLS \cite{bodenreider2004unified} as these resources contain well-defined types and relations between words and concepts. Recent works have shown that deep learning models are more effective on this task by learning linguistic knowledge from large-scale text data and representing text (words and sentences) as continuous trainable embedding vectors \cite{hochreiter1997long, devlin2018bert}. 

Among these works, deep Transformer-based models such as the Bidirectional Encoder Representations from Transformers (BERT) has shown very promising performance, thanks to the contextualized word representations learned through its multi-head attention mechanism and unsupervised pre-training on large corpus \cite{vaswani2017attention}. To further instill knowledge into BERT, recent works have proposed different training tasks beyond the original masked language modeling and next-sentence prediction tasks \cite{devlin2018bert}. These include other unsupervised pre-training tasks such as span prediction \cite{joshi2019spanbert} and domain-specific pre-training \cite{gururangan2020dont}, as well as knowledge-based tasks such as semantic role prediction \cite{zhang2019semantics}, entity recognition \cite{zhang2019ernie}, and relation prediction \cite{weijie2019kbert}.
Among these approaches, some are able to improve the semantic textual similarity (STS) task performance \cite{joshi2019spanbert}, while others are sometimes detrimental to the task \cite{zhang2019ernie, zhang2019semantics}.

In this paper, we explore a different approach to incorporating external knowledge into deep Transformers for STS tasks. Instead of creating additional training tasks, we directly inject knowledge  into a Transformer's multi-head attention.
On the one hand, deep Transformers are usually seen as complex ``black boxes'' that can only be improved through well-formulated training (or fine-tuning) tasks. On the other hand, the research community has recently seen a surge of interest in ``opening the black box'' to understand the internal mechanism of Transformer-based models. These include analyses of  layer-wise attentions, syntactic knowledge, semantic knowledge, and world knowledge learned at each layer  \cite{clark2019what, hewitt2019structural,tenney2019you,ettinger2020bert}. We refer the reader to Rogers et al. \cite{rogers2020primer} for a comprehensive synthesis. Motivated by these recent studies, we aim to take one step further -- in addition to \emph{observing} the mechanisms inside BERT, we explore the possibility of purposefully \emph{steering} its internal mechanism and study the effects of such intervention. In particular, we use prior knowledge to guide the attention of BERT towards a better performance on STS tasks. Although previous work has proposed to use prior knowledge in guiding the attention of other deep learning models, such as recurrent neural networks \cite{chen2016knowledge} and long short-term memory networks \cite{chen2019deep}, to the best of our knowledge, we are the first to explore direct intervention of BERT's attention mechanism.

To make informed decisions on what kind of knowledge to add and where to add it in BERT, we conducted in-depth analyses of the BERT model on the STS task. The results inform us to inject word similarity knowledge into BERT's attention at the first layer. Extensive experiments on standard STS datasets demonstrate that our approach  consistently improves BERT on STS tasks, and the improvement effect is most salient when training data size is very limited. 
The main contributions of this paper are as follows:
\begin{itemize}
\item We show that while a pre-trained BERT model has prior knowledge on STS tasks which a non-Transformer deep model does not possess, its knowledge in word similarity is still inadequate and can be further improved.

\item We propose an efficient and effective method for injecting word similarity knowledge into BERT -- not through adding another training task, but by directly guiding the model's attention. This approach not only performs at least as well as adding a new training task, but also saves substantial training time.

\item We show that the proposed method is able to consistently improve the STS performance of BERT and the benefit is especially salient when training data is scarce.
\end{itemize}

\section{Related Work}
Semantic textual similarity is a well-studied task in natural language processing (NLP). In most previous works, feature engineering was the main approach. Early research on this task explored different types of sparse features and confirmed their value. This includes (1) syntactical and lexical features extracted from word and sentence pairs \cite{das2009paraphrase, xu2014extracting},  (2) knowledge-based features using WordNet, which make extensive use of word similarity information \cite{fernando2008semantic}, (3) semantic relation knowledge \cite{iftene2007hypothesis} and logical rules \cite{beltagy2015representing} derived from WordNet, (4) word-alignment features which discover and align similar semantic units in a pair of sentences \cite{sultan2014back, sultan2015dls}.
Before the advent of BERT, other types of deep neural networks have already been used to solve STS tasks. These include feedforword neural networks (FNN), convolutional neural network s (CNN) and long short-term memory-Networks (LSTM). For example, Hu et al. used a CNN model that combines hierarchical structures with layer-by-layer composition and pooling \cite{hu2014convolutional}. Yin et al. presented a general attention-based CNN for modeling a pair of sentences \cite{yin2016abcnn}. Parikh et al.  used attention to decompose the problem into sub-problems that can be solved separately in the form of natural language inference (NLI) tasks \cite{parikh2016decomposable}. Tomar et al. proposed a variant of the decomposable attention model which achieved good results on paraphrase identification task \cite{tomar2017neural}. Besides, Chen et al. utilized tree-LSTM and soft-alignment to improve the performance of the ESIM model on NLI tasks \cite{chen2017enhanced}.  Lan et al. showed that the tree-LSTM model also did well in STS task  \cite{lan2018neural}.   Among many variants of the ESIM  \cite{chen2018neural, li2018attention}, KIM  leveraged external knowledge to improve performance of ESIM on semantic similarity tasks \cite{chen2018neural}.

In recent years, the shift from neural network architecture engineering to large-scale pre-training has significantly improved NLP tasks, demonstrating the power of unsupervised pre-training.
Outstanding examples include Embedding from Language Models (ELMo) \cite{peters2018deep}, Generative Pre-trained Transformers  (GPT) \cite{radford2018improving},  Bidirectional Encoder Representations from Transformers (BERT) \cite{devlin2018bert}, and Generalized Auto-regressive Pre-training (XLNet) \cite{yang2019xlnet}. Providing fine-grained contextual word embedding, these pre-trained models can be either easily applied to downstream tasks as encoders or directly fine-tuned for downstream tasks. As the most prominent model in recent years, BERT and many of its variants, including AlBERT \cite{lan2019albert}, RoBERTa \cite{liu2019roberta}, SemBERT \cite{zhang2019semantics}, ERNIE \cite{zhang2019ernie}, K-BERT \cite{weijie2019kbert}, and DeBERTa \cite{he2020deberta}, have achieved superior results in many NLP tasks. Among them, SemBERT, ENRIE and K-BERT all add knowledge to the orginal BERT model, but in different ways. SemBERT and K-BERT are fine-tuning methods without pre-training  and they are capable of loading model parameters from a pre-trained BERT. SemBERT incorporates explicit contextual semantics from pre-trained semantic role labeling. K-BERT is a knowledge-enhanced language model, in which knowledge triples are injected into the sentences as domain knowledge. K-BERT is only trained on Chinese text corpus. ENRIE is an improvement on top of BERT that utilizes both large-scale textual corpora and knowledge graph. It takes advantage of lexical, syntactic, and semantic knowledge information simultaneously. Although the above three BERT variants are shown to improve performance on a variety of language understanding tasks, the results largely depend on the size of task-specific training data. For instance, the performance of ERNIE was reported to be unstable on semantic similarity tasks with small datasets \cite{zhang2019ernie}.

\section{What has BERT already known about semantic textual similarity?}
\label{sec:bertanalysis}

Before adding task-specific knowledge to BERT, the first and foremost question is:  
\emph{what has BERT already known (and not known) about the task?} Answering this question will allow us to design our approach in an informed, targeted manner.

To operationalize our answer to the above question, we use a classical STS dataset -- the Microsoft Research Paraphrase Corpus (MPRC) \cite{dolan2005automatically}, as a pilot data set. Each data instance is a pair of sentences. The goal is to detect if one is a paraphrase of the other (binary classification).  We conduct two pilot studies as follows.

\textbf{(1) Data Augmentation Study}.
In this study, we augment the pilot dataset with various types of prior knowledge that are potential useful for determining semantic textual similarity, and compare the performance of BERT trained on the original data vs. the augmented data. The intuition is that if a particular data augmentation strategy can improve BERT's STS performance, it indicates that BERT still lacks the corresponding knowledge. Otherwise it implies that BERT has already ``known'' the corresponding  knowledge. As a comparison, we apply the same procedure to a non-Transformer model for the STS task, the Enhanced Sequential Inference Model (ESIM) \cite{chen2017enhanced}, and see if the same data augmentation strategy can benefit such a model (which indicates that the model lacks the corresponding knowledge). We defer detailed descriptions of BERT and ESIM to Section \ref{sec:method}.

\textbf{(2) Layer-wise Performance Study}.
In this study, we freeze a BERT model (except for the softmax classification output head) and use the pretrained contextualized word representation to perform the STS task. The goal is to observe \emph{where} BERT stores the most  STS-related knowledge, and where such knowledge is the most lacking. This allows us to make an informed decision on which BERT layer(s) to inject prior knowledge.

Below we discuss the results and implications of the two studies.

\subsection{Effect of Data Augmentation}
\label{sec:data-aug}

In short, data augmentation techniques use prior knowledge and relatively simple algorithms to derive new training data from  original training data  \cite{xie2019unsupervised}. It is an effective way of enriching the original data with task-specific knowledge. Below we describe several data augmentation strategies and the corresponding prior knowledge related to STS. 
\begin{enumerate} [(1)]
    \item \emph{Split and swap}: For each sentence, split it at a random position and swap the two segments. The assumption is that although a sentence could be ungrammatical after such an operation, its essential meaning should be preserved.
    \item \emph{Add random word}: For each sentence, pick a random word from the vocabulary that is not in the sentence, and insert it at a random position in that sentence. The assumption is that  such an out-of-context word acts as a noisy typo and should not affect the main meaning of the sentence.
    \item \emph{Back translation}: All sentences in MRPC are in English. Using Google Translate, we translate the sentence to Chinese and then back to English. The assumption is that translation should rewrite the sentence in a different way but preserve meaning.
    \item \emph{Add high-TfIdf word}: For each sentence, find the word with the highest TfIdf weight and insert it at a random position in that sentence. Words with high-TfIdf weights are usually content words, therefore repeating them should not change the meaning of the sentence by much.
    \item \emph{Delete low-TfIdf word}: Find $k$ words with the lowest TfIdf weights in the sentence pair. Then for each of these words, delete it from the sentence with probability $p$. We set $k=5, p = 0.5$. Words with low TfIdf weights are usually \emph{stop words} or \emph{functional words} (e.g., ``the'', ``of'', ``and''), and therefore may not be essential to the meaning of the sentence.
    \item \emph{Replace synonyms}: For each word in a sentence, see if the word has synonyms in WordNet. If so, pick the first word in the synonym list that is not the word itself. The assumption is that replacing words with their synonyms does not change the meaning of a sentence. In principle, other resources such as ConceptNet and UMLS also provide synonym knowledge and can be used here in place of WordNet.
\end{enumerate}

\begin{figure}[htpb!]
  \centering
  \includegraphics[width=.8\linewidth]{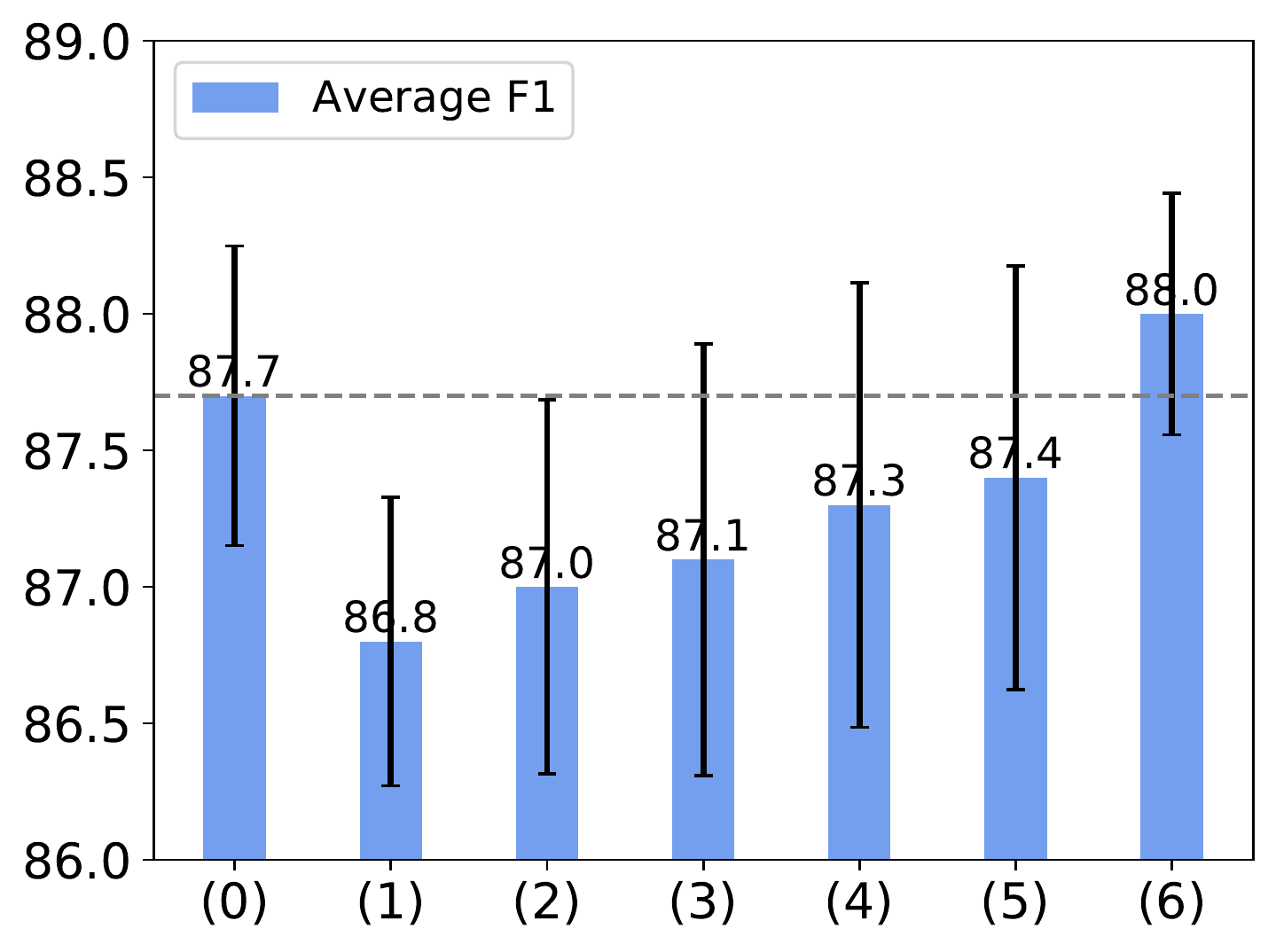}
  \caption{Effect of augmenting the MRPC data set for BERT. The X-axis represents different data augmentation strategies corresponding to the order in the text, where (0) represents the original training data without augmentation, and the serial number (1) to (6) represents \emph{Split and swap},  \emph{Add random word},  \emph{Back translation},  \emph{Add high-TfIdf word},  \emph{Delete low-TfIdf word},  \emph{Replace synonyms}, respectively. Error bars show $\pm 1$ standard deviation around the average of 10 runs.}
  \label{fig:data-aug-bert}
\end{figure}

We use the standard train-test split provided in the MRPC dataset. Then, we apply each of the above data augmentation strategies on the training data, each resulting in a 2$\times$ increase of training data size. Figure \ref{fig:data-aug-bert} shows the $F_1$ performance of BERT trained by each augmented dataset, as well as the performance without data augmentation. To minimize the effect of randomness in BERT training, performance levels are averaged across 10 different runs.

The results show that except for \emph{Replace synonyms}, all other strategies lead to a performance drop. This indicates that BERT may have already understood the importance of stop words and content words in STS tasks, and it knows to use syntactic ordering and semantic coherence when inferring semantic similarity.  The fact that back translation does not help implies that back-translated sentences may contain semantic shifts and/or syntactic errors, which can mislead BERT.

\begin{figure}[htpb!]
  \centering
  \includegraphics[width=.8\linewidth]{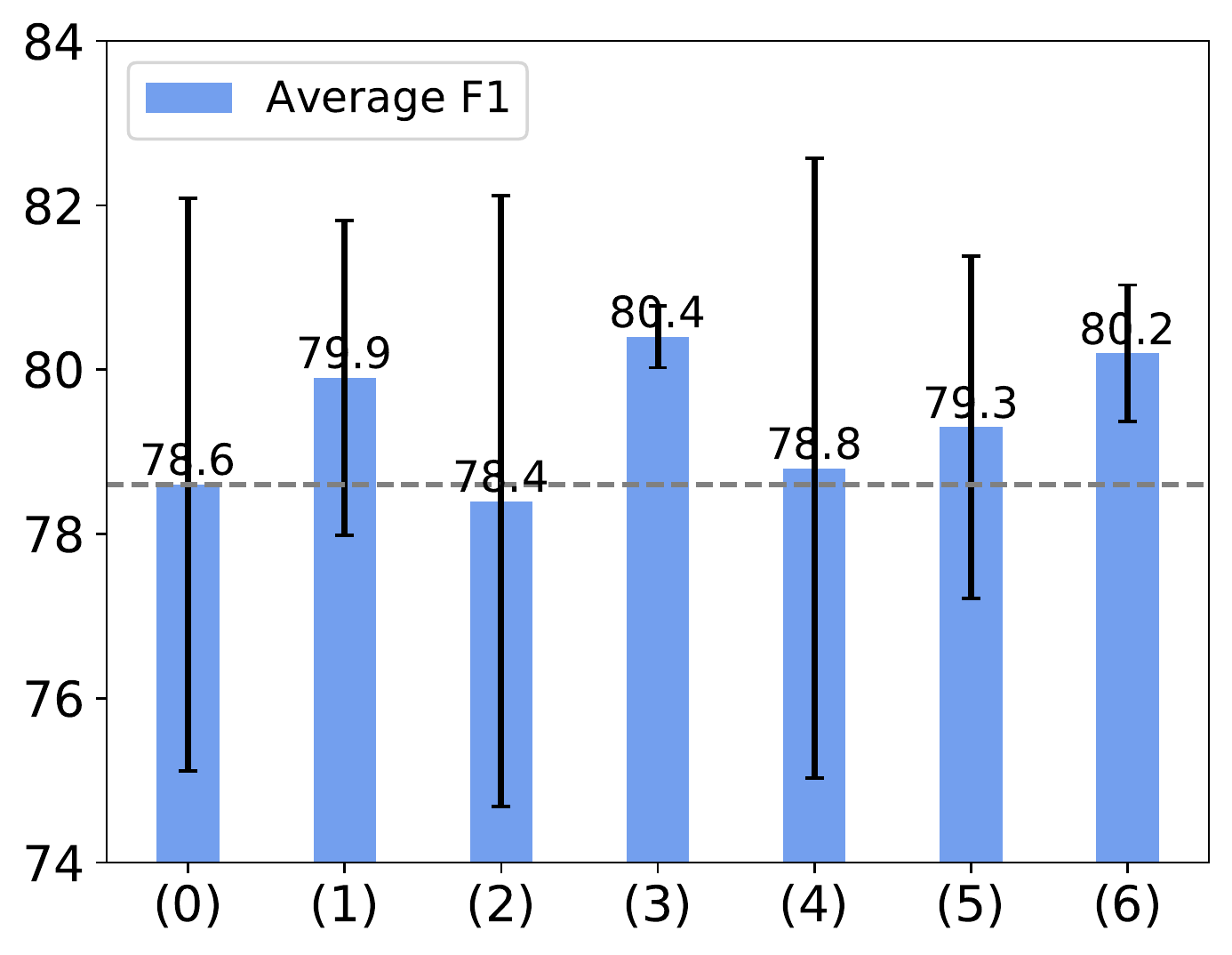}
  \caption{Effect of augmenting the MRPC data set for ESIM. X-axis labels are defined similarly as those in Figure \ref{fig:data-aug-bert}.}
  \label{fig:data-aug-esim}
\end{figure}

As a comparison, we apply the same augmented data to ESIM. The results are shown in Figure \ref{fig:data-aug-esim}. Note that BERT  outperforms ESIM by a large margin. Here we focus on the relative differences made by different augmentation strategies on ESIM.
ESIM benefits from all data augmentation strategies except for \emph{Add random word}. Both \emph{Back translation} and \emph{Split and swap} are effective strategies (>1\% absolute gain in $F_1$), which do not apply for BERT. 
This indicates that a pretrained BERT model has rich prior knowledge for the STS task that a pre-Transformer model does not possess. 
Similar to BERT, \emph{Replace synonyms} can substantially benefit ESIM as well.

\subsection{Different BERT Layers on the STS Task}

\begin{figure}[t!]
  \centering
  \includegraphics[width=.8\linewidth]{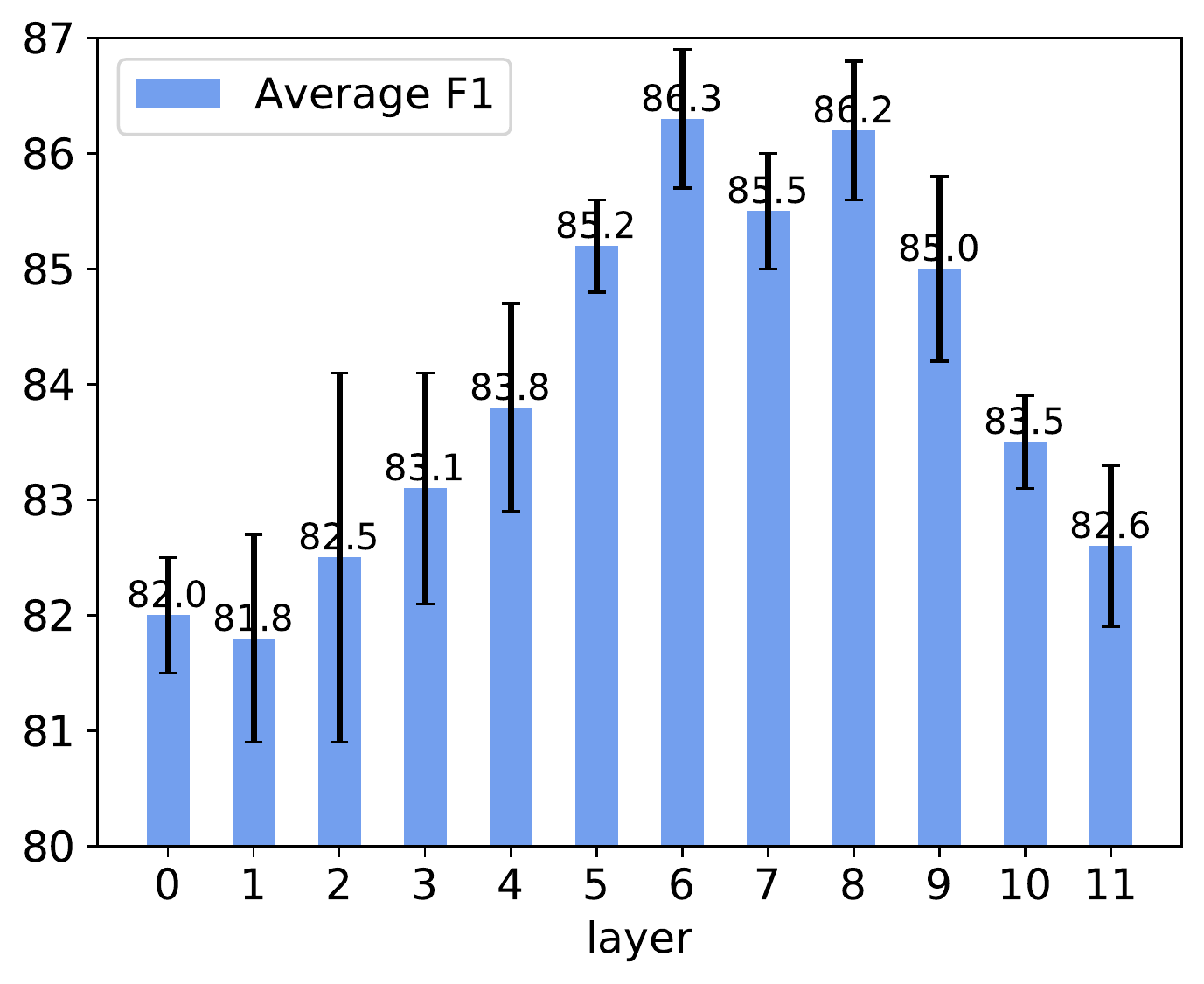}
  \caption{Performance of each BERT layer on MRPC data set.}
  \label{fig:bert-layer-analysis}
\end{figure}

Figure \ref{fig:bert-layer-analysis} shows the performance of using each pre-trained BERT layer on the MRPC paraphrase identification task. In this study, the original  training data was used without any augmentation. Again, the performance levels are averaged across 10 different runs.

Since the BERT layer parameters are ``frozen'' (not trainable) in this task, the performance shows that BERT's middle layers are inherently good at this task, while lower and upper layers are not. This echoes with recent findings on BERT layers. Lower layers were found to perform broad attention across all pairs of words \cite{clark2019what}. Middle layers were found to mostly capture transferable syntactic and semantic knowledge \cite{hewitt2019structural,tenney2019bert}. It is not surprising that the upper layers do not perform well, as these layers are specifically tuned towards the pre-training tasks of BERT -- masked language modeling and next-sentence prediction -- not STS tasks.

\vspace{.1in}
To summarize, the data augmentation study  shows that BERT knows about the importance of stop words, content words, and syntactic structure for the STS task, but its knowledge in word synonyms can be further improved.
The layer-wise performance study shows that lower layers of BERT needs the most improvement for STS tasks. Combining results from the two studies, we identify a promising direction, i.e., to incorporate synonym knowledge into low layers of BERT (next section). We would like to note that the above analysis and reasoning procedure can be potentially applied to identify ``knowledge deficiency'' of BERT in other NLP tasks as long as one can associate different kinds of knowledge with different subsets of training data.

\section{Proposed Algorithm }
\label{sec:method}

In this section, we design a general algorithm for incorporating synonym knowledge into the BERT model. According to the findings in Section \ref{sec:bertanalysis}, this knowledge should be added to lower layers of BERT. Since each attention layer computes similarity between all pairs of words, and it has been  shown that the attention in the first layer is broad and uninformed \cite{clark2019what, rogers2020primer}, we decide to use word similarity knowledge to modulate the attention at the first layer.  As a comparison baseline, We add the same knowledge into the ESIM model. 
Below we describe our approach in detail.

\subsection {Word Similarity Matrix}
\label{sec:sim-matrix}

Given two pieces of text (i.e., sentences) $a = (w_1, \cdots, w_a, \cdots, w_{l_a})$ and $b = (w_1, \cdots, w_b, \cdots, w_{l_b})$, we construct a word similarity matrix $S$ of size $l_a \times l_b$.
The goal of this matrix $S$ is to increase BERT's attention on semantically similar word pairs. 

We calculate the value of each cell in $S$ based on semantic relations in WordNet. For a lexical pair $(w_{a}, w_{b})$, if words in the pair are synonyms in WordNet, the cell value $S_{ab} = 1$. Otherwise, if $w_{a}$ and $w_{b}$ are not synonyms, we set $S_{ab}$ as a value in $[0,1]$ according to the method proposed by Wu-Palmer \cite{wu1994verbs}, which calculates word similarity based on their topological distance in WordNet. 
In cases where one or both words in a pair cannot be found in WordNet, or the pair of words do not have a valid Wu-Palmer similarity value (e.g. if one word is a stop word like ``into''), 
we directly set the pairwise similarity value to $0$. For proper names of people and places, if words in the pair are the same, the value will be set to $1$, otherwise it will be $0$. Figure 4 visualizes the heat map of the similarity matrix constructed for two sentence pairs. We compute word similarity matrices for all sentence pairs in both training and test dataset as a preprocessing step.

Note that our method is general and compatible with other semantic similarity resources, such as Wikipedia, ConceptNet, and UMLS (Unified Medical Language System) Metathesaurus. We can construct $S$ using  word/concept similarity knowledge in these resources as well. Also note that simple word lookup does not resolve word sense ambiguity, and WordNet does not cover many acronyms and abbreviations. We leave further refinements of word similarity matrix $S$ for  future work.

\begin{figure}[h]
  \centering
  \includegraphics[width=\linewidth]{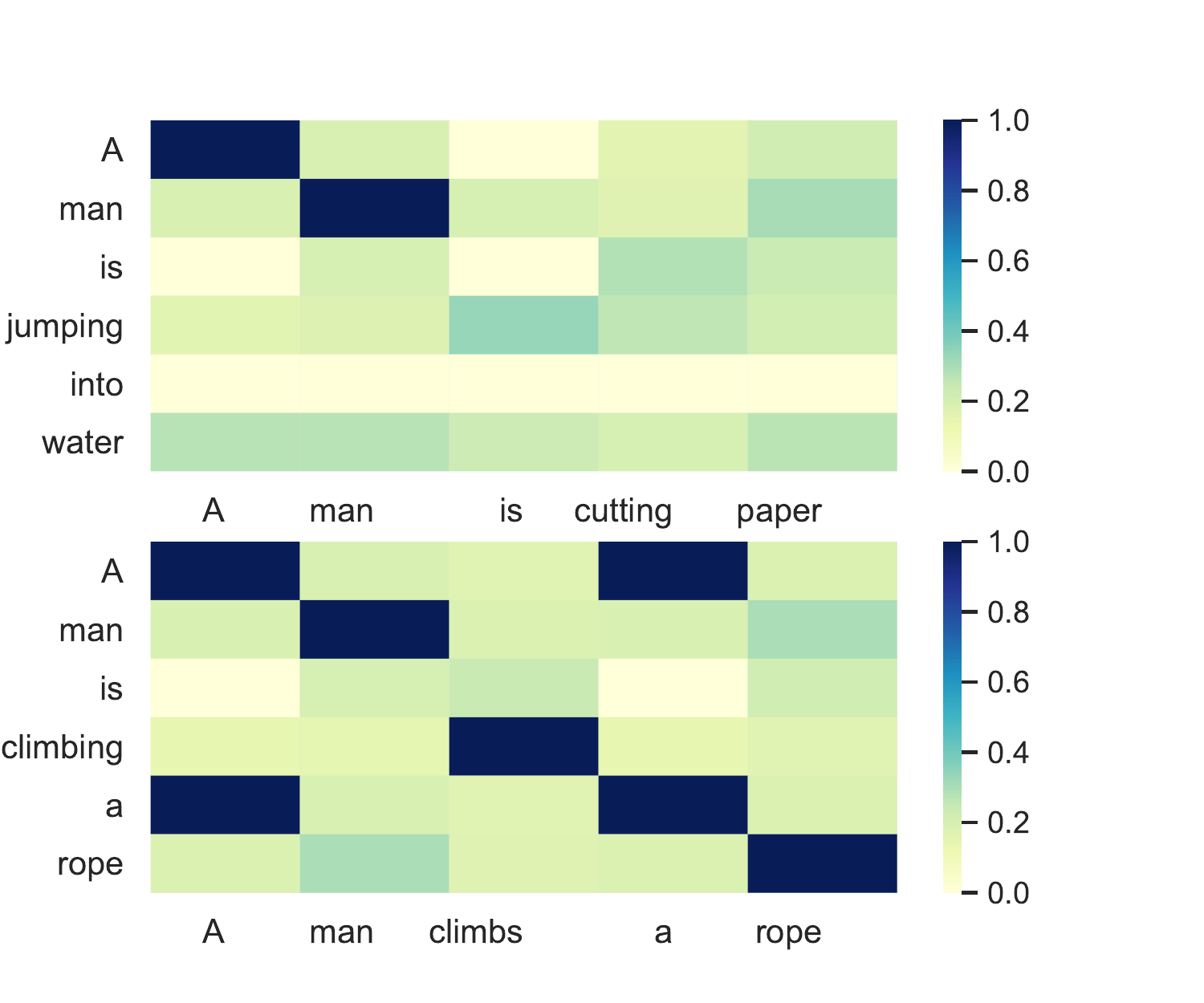}
  \caption{Heat map for two sentence pairs taken from the STS-B dataset, where the ground-truth similarity score of the first sentence pair (``A man is jumping into water'' and ``A man is cutting paper'') is labeled as 0.8, and the similarity score of the second sentence pair is 5 (``A man is climbing a rope'' and ``A man climbs a rope''). The higher score, the more similar the two sentences.}
  \end{figure}

\subsection{Bidirectional Encoder Representations from Transformers (BERT)}
BERT is a pre-trained language model with state-of-the-art performance on many NLP tasks. We fine-tune Google's pre-trained BERT-base model in our implementation. 

Word similarity knowledge is added into BERT's multi-head attention (both self-attention and cross-attention). 
In the embedding stage, we use the summation of three parts: token embedding, position embedding, and segment embedding, which is the same as BERT. In the Transformer stage, our method is similar to BERT.

\subsubsection{Multi-Head Attention in BERT}

BERT's attention function can be described as a  mapping from query vector $Q$ and a set of key-value vector pairs $(K, V)$ to an output vector -- the attention strengths. Multi-head attention linearly projects the queries, keys and values $h$ times ($h$ is the number of ``heads'') with different linear projections to $d_k$, $d_k$, and $d_v$ dimensions, respectively. Next, on each head of these projections of queries, keys and values, it performs the attention function in parallel. This produces $d_v$-dimensional output values. These values are concatenated and projected again to get the final attention:
\begin{equation}
\begin{split}
Multi&Head(Q, K, V) = Concat(head_{1},\dots,head_{h})W^{O} \ ; \\
&head_{i} = Attention(QW_{i}^{Q}, KW_{i}^{K}, VW_{i}^{V})\ ,
\centering
\end{split}
\end{equation}
where $W_{i}^{Q} \in \mathbb{R}^{d_{model}\times d_{k}}$, $W_{i}^{K} \in \mathbb{R}^{d_{model}\times d_{k}}$,$W_{i}^{V} \in \mathbb{R}^{d_{model}\times d_{v}}$ and $W_{i}^{O} \in \mathbb{R}^{hd_{v}\times d_{model}}$ are parameter matrices representing the projections. $d_{model}$ is BERT's hidden layer size. The calculation of BERT's attention uses scaled dot-product:
\begin{equation}
\begin{split}
&scores = QK^{T} + MASK \ ;\\
Atten&tion(Q, K, V) = softmax(\frac{scores}{\sqrt{d_{k}}})V \ ,
\centering
\end{split}
\end{equation}
where $MASK$ is a matrix used in masked language modeling, the pretraining task for BERT.

\subsubsection{Knowledge-Guided Attention}

\begin{figure}[h]
  \centering
  \includegraphics[width=\linewidth]{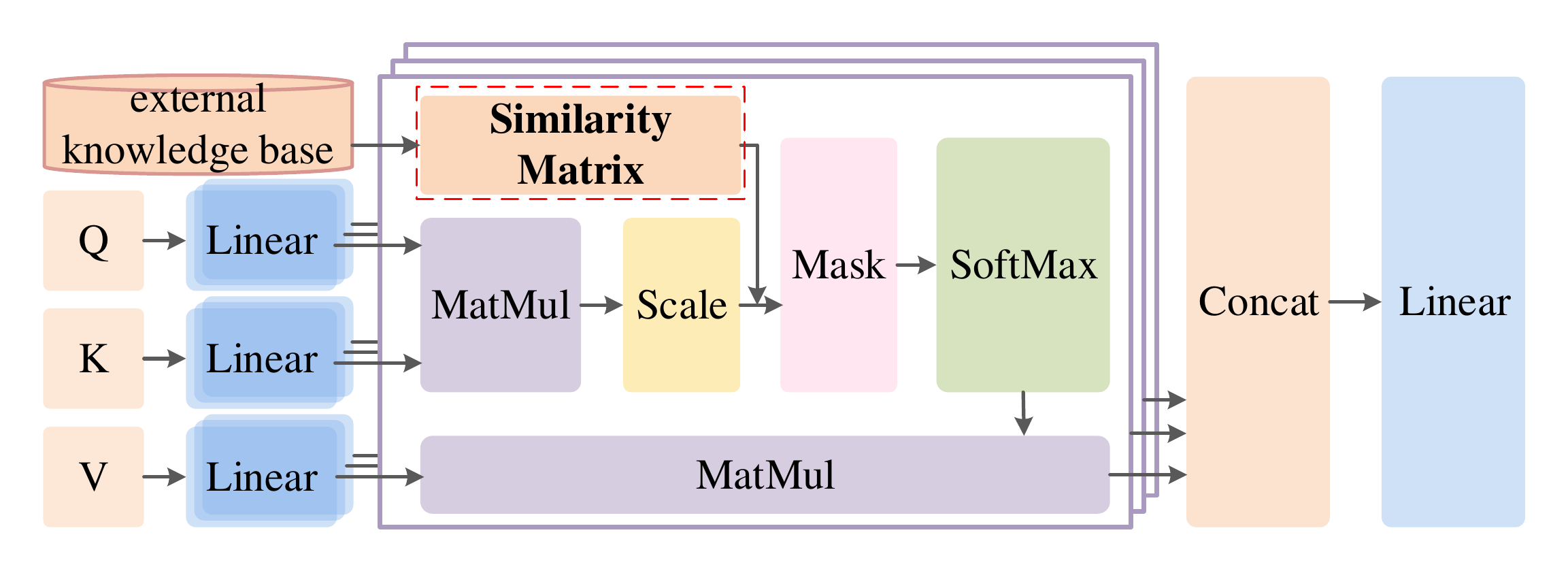}
  \caption{Injecting word similarity knowledge in BERT's multi-head attention.}
  \label{fig:bert-sim}  
\end{figure}

To add prior knowledge, we make adjustments to the multi-head attention phase as shown in Figure \ref{fig:bert-sim}. We calculate the Hadamard product (i.e., element-wise product) of the $scores$ using the similarity matrix $S$ to make the model pay more attention to the word pairs with higher similarities in two sentences:
\begin{equation}
\begin{split}
&scores = QK^{T} \odot S + MASK \ ;\\
Atten&tion(Q, K, V) = softmax(\frac{scores}{\sqrt{d_{k}}})V \ ,
\centering
\end{split}
\end{equation}
where $S$ represents the word similarity matrix we calculated in advance (Section \ref{sec:sim-matrix}). As mentioned before, we only add such prior knowledge in the attention of the first BERT layer, as that layer's attention is broad and uninformed \cite{clark2019what,rogers2020primer}. 

\subsection {Enhanced Sequential Inference Model }

In this section, we show that our method can also be used to incorporate knowledge into the Enhanced Sequential Inference Model (ESIM). This model explicitly computes pairwise word similarities as a component \cite{chen2017enhanced}. Before  BERT was proposed, ESIM was one of the state-of-the-art methods for  sentence pair modeling tasks. 
In this paper, we use ESIM as a non-Transformer model for baseline comparison.

\subsubsection{The Original ESIM}
ESIM first uses a bi-directional long short-term memory network (BiLSTM) to encode the input sentences $a$ and $b$. For the $i$-th token of sentence $a$, the hidden state vector generated by the BiLSTM is denoted as $\overline{a}_{i}$. Similarly we can define a state vector  $\overline{b}_{j}$ for the $j$-th token of sentence $b$:
\begin{equation}
\begin{split}
\overline{a}_{i} = BiLSTM(a,i)\quad \forall_{i} \in [1,\dots,l_{a}] \ ;\\
\overline{b}_{j} = BiLSTM(b,j)\quad \forall_{j} \in [1,\dots,l_{b}] \ .
\centering
\end{split}
\end{equation}

ESIM computes the attention weights $e_{ij}$ as the inner product of a hidden state pair $\langle \overline{a}_{i}, \overline{b}_{j}\rangle $ for all tokens in sequences $a$ and $b$:
\begin{equation}
e_{ij}= \overline{a}_{i}^{T}\overline{b}_{j} \ .
\centering
\end{equation}

The attention weights are then normalized and used as coefficients to compute a ``soft alignment'' between a word in one sentence and all words in the other sentence:
\begin{equation}
\begin{split}
\overline{\beta}_{i}&=\sum_{j=1}^{l_b}\frac{\exp(e_{ij})}{\sum_{k=1}^{l_a} \exp(e_{ik})} \overline{b}_{j} \ ; \\
\overline{\alpha}_{j}&=\sum_{i=1}^{l_a}\frac{\exp(e_{ij})}{\sum_{k=1}^{l_b} \exp(e_{kj})} \overline{a}_{i} \ ,
\centering
\end{split}
\end{equation}
where $\overline{\beta}_{i}$ is the ``subphrase'' in $b$ that is softly aligned to words in ${a}$ and vice versa for $\overline{\alpha}_{j}$.

After soft-alignment, ESIM uses Tree-LSTM model to help collect local inference information over linguistic phrases and clauses. The output of the Tree-LSTM model is then max-pooled, averaged, and concatenated as input to a final multilayer perception.  We refer the reader to \cite{chen2017enhanced} for more details.

\subsubsection{Knowledge-Guided ESIM}
\begin{figure}[htbp]
  \centering
  \includegraphics[width=1\linewidth]{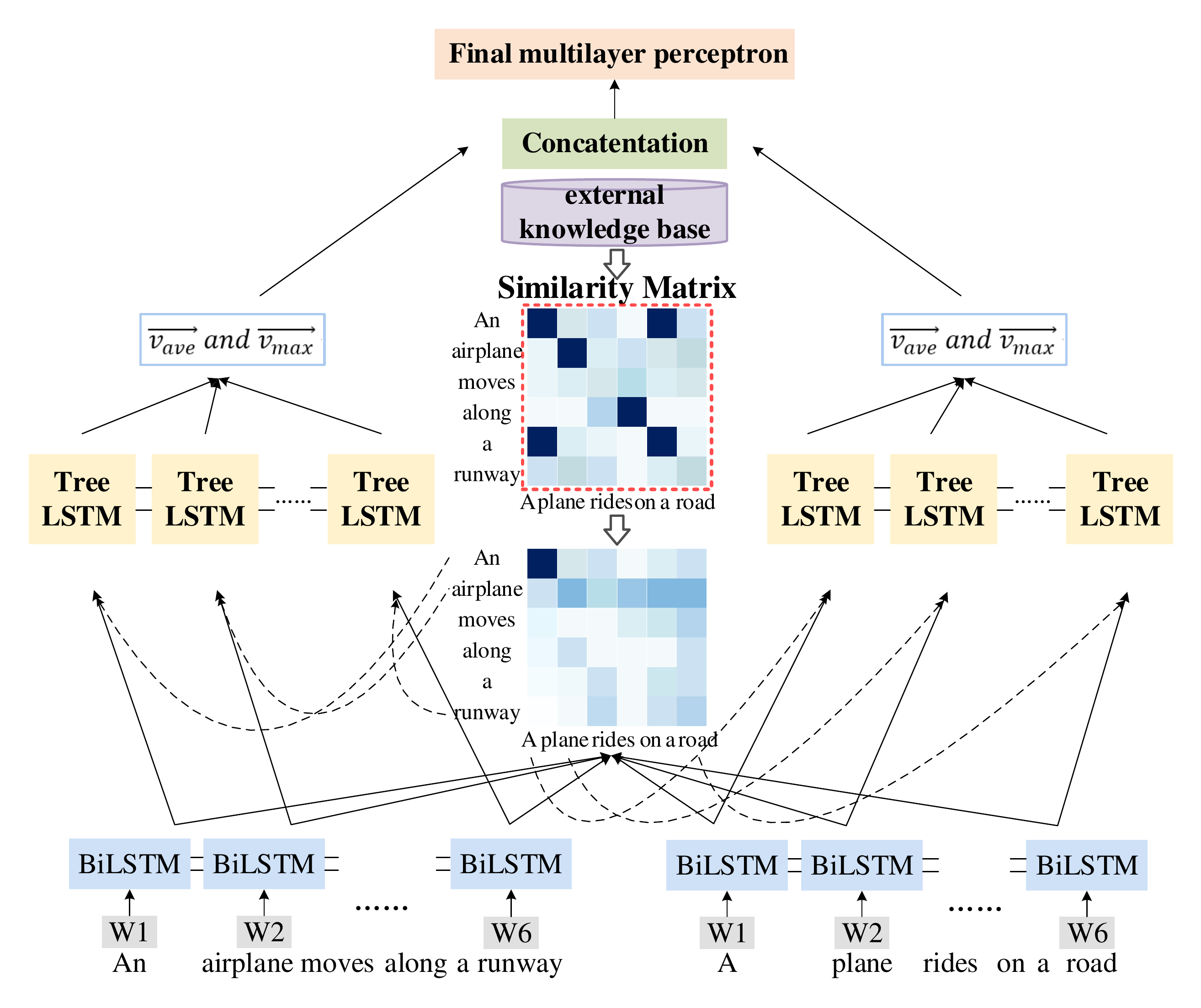}
  \caption{Injecting Word Similarity Knowledge in ESIM}
  \label{fig:esim}
  \end{figure}
 The overall architecture of the knowledge-enhanced ESIM is illustrated in Figure \ref{fig:esim}.
To inject word similarity knowledge, we also compute the Hadamard product between the attention weight matrix $e$ and the word similarity matrix $S$ defined in Section \ref{sec:sim-matrix}:
\begin{equation}
p_{ij} = e_{ij}\odot S_{ij} \ .
\centering
\end{equation}

So that the soft alignment will be updated as follows:
\begin{equation}
\begin{split}
\overline{\beta}_{i}&=\sum_{j=1}^{l_b}\frac{\exp(p_{ij})}{\sum_{k=1}^{l_a} \exp(p_{ik})} \overline{b}_{j} \ ; \\
\overline{\alpha}_{j}&=\sum_{i=1}^{l_a}\frac{\exp(p_{ij})}{\sum_{k=1}^{l_b} \exp(p_{kj})} \overline{a}_{i} \ ,
\centering
\end{split}
\end{equation}
The remaining steps are similar to the original ESIM.

\section{Experiments}
In this section, we evaluate our proposed method and compare it with baseline models that do not incorporate prior knowledge.

The benefit of incorporating prior knowledge in machine learning models is most salient when training data is small. Indeed, when training data is abundant, those data already contain sufficient task-specific knowledge, diminishing the benefit of prior knowledge.  Therefore besides the standard setting where different models are evaluated on 100\% training data, we are more interested in evaluating our approach as we vary the size of training data from small to large. This motivates us to further conduct learning curve analyes. 
\begin{table*}[t]
\caption{Semantic textual matching performance in four  datasets (\%), based on our re-implementation of each method in PyTorch. Each performance value of ESIM and ESIM-Sim (BERT$_\text{base}$ and BERT$_\text{base}$-Sim) are the average of 10 (5) runs with different random seeds, respectively. Values in the parentheses are standard deviations (SD). PC: Pearson correlation coefficient. SC: Spearman's rank correlation coefficient. The best average performance in each column is in bold.   } 
\renewcommand\tabcolsep{4.0pt}
	\centering
	\begin{spacing}{1}
	\begin{tabular}{l|p{1.2cm}<{\centering}p{2cm}<{\centering}|p{1.6cm}<{\centering}p{1.6cm}<{\centering}|p{1.2cm}<{\centering}p{2cm}<{\centering}|p{1.2cm}<{\centering}p{2cm}<{\centering}}
	\toprule
	&\multicolumn{2}{c|}{\textbf{MRPC}} &\multicolumn{2}{c|}{\textbf{STS-B}}
	 &\multicolumn{2}{c|}{\textbf{QQP}}
	 &\multicolumn{2}{c}{\textbf{TwitterURL}}\\

	& F1 (SD) & Macro-F1 (SD)	 & PC (SD) & SC (SD) & F1 (SD) & Accuracy (SD) & F1 (SD) & Macro-F1 (SD)  \\\hline
		
		ESIM     & 78.5 (2.0)  &56.5 (6.2)  & 47.2 (0.7)  & 44.0 (0.8)  &87.6 (0.5)  & 87.5 (0.5)  &59.1 (2.3) &71.3 (2.1)\\
		ESIM-Sim & 79.1 (1.5)  &60.3 (5.8)  & 59.3 (1.3)  & 56.4 (1.4)  &87.5 (0.5)  & 87.6 (0.5)  &65.3 (1.7) &78.2 (1.3)\\ 
		$\text{BERT}_\text{base}$      & 87.0 (1.5) &80.7 (1.7)   & 84.4 (0.7) &82.9 (0.7)   & \textbf{90.7} (0.4) & \textbf{90.8} (0.4) &76.0 (0.7)  & 84.8 (0.5)\\ 
		$\text{BERT}_\text{base}$-Sim  & \textbf{88.3} (0.3) & \textbf{81.2} (1.1)   & \textbf{85.1} (1.2) & \textbf{83.8} (1.1)   & 90.5 (0.4) &90.7 (0.4) & \textbf{76.2} (0.7)  & \textbf{85.0} (0.6)\\
	\bottomrule
	\end{tabular}
	\label{tb:benchmark_stat}
	\end{spacing}

\end{table*}

\subsection{Datasets}
We use four popular benchmark datasets in our experiments: two relatively small semantic textual similarity datasets and two relatively large paraphrase identification datasets. 
\begin{itemize}
\item \textbf{MRPC} \cite{dolan2005automatically} is a corpus of sentence pairs with artificial annotations automatically extracted from online news sources to indicate whether each pair of sentences captures the paraphrase/semantic equivalence relationship through binary judgment. It has 4,076 train data and 1,725 test data. In this paper, we split 10\% training data as the validation set according to GLUE \cite{wang2019glue} standardized splits. In Section \ref{sec:bertanalysis}, we used this dataset as a pilot to understand BERT.
\item \textbf{STS-B} \cite{cer2017semeval} is a collection of sentence pairs extracted from news headlines, video headlines, image headlines, and natural language inference data. It comprises a selection of the English datasets used in the STS tasks organized in the context of SemEval between 2012 and 2017 which includes 5,749 train data, 1,500 development data and 1,379 test data.
\item \textbf{QQP} \cite{iyer2017first} is a dataset used to determine whether a question pair is duplicated. It consists of more than 400,000 rows of potential question duplicate pairs collected from Quora.com. In this paper, we use the same data and split by Wang in article \cite{wang2017bilateral}, with 10,000 question pairs each for development and test. Besides, in both development and test set, the number of both paraphrasing and non-paraphrasing sentence pairs is 5000. Accuracy is used as the evaluation metric for this dataset.
\item \textbf{Twitter-URL} \cite{lan2017continuously} is collected from tweets that share the same URL of news articles by Lan. It includes 56,787 sentence pairs, and each sentence pair is annotated by 6 Amazon Mechanical Turk workers. Therefore, a total of 6 workers judged if this pair is a paraphrase or not. If n $\le$ 2 workers were positive, we treat them as non-paraphrasing; if n $\ge$ 4, we treat them as paraphrasing; if n $=$ 3, we discard them. After this treatment, there were 42,220 pairs for training and 9,334 pairs for test.
\end{itemize}

\subsection{Implementation Details}
We implement the models with the same PyTorch framework. Below, we summarize the implementation details that are key for reproducing results for each model. (The source code is available at \url{https://github.com/xiatingyu/Bert_sim}).

\begin{itemize}
\item \textbf{BERT}: We compare our proposed model with a BERT model without prior knowledge. Both models adopt the configuration of Google's BERT$_\text{base}$ \cite{devlin2018bert}. We set the number of both self-attention layers and heads as 12, and the  dimension of embedding vectors as 768. The total number of trainable parameters of both the original BERT$_\text{base}$ and our proposed model (called BERT$_\text{base}$-Sim) are the same (110M), therefore we are performing a head-to-head comparison.
\end{itemize}
\begin{itemize}
\item \textbf{ESIM}: Word embeddings of ESIM model is initialized with 840B-300d Glove  word vectors \cite{pennington2014glove}. Embeddings of out-of vocabulary words are randomly initialized. All parameters, including word embeddings, are updated during training. In order to verify the influence of prior knowledge on the model effect, we use the same parameters mentioned in paper \cite{chen2017enhanced}, including optimization method, learning rate, dropout rate. We call our proposed model ESIM-Sim.
\end{itemize}

\subsection{Performance Across Four STS Datasets}
 \begin{figure*}[t]
  \centering
  \subfigure[MRPC]
  {
  \begin{minipage}{.23\textwidth}
    \centering
    {\includegraphics[width=\linewidth]{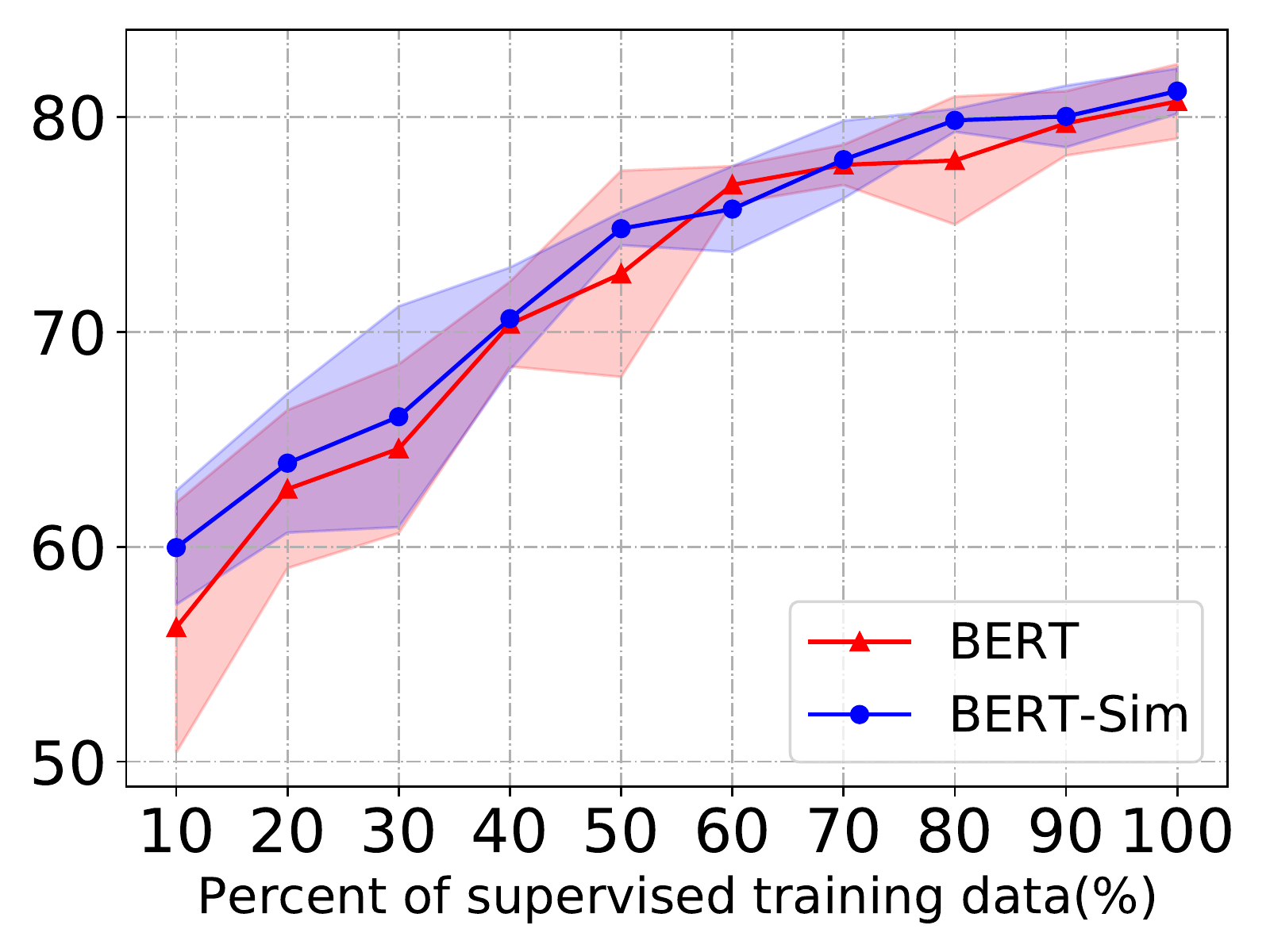}}
    \label{fig:illustrate_cases}
  \end{minipage}
  }
  \subfigure[STS-B]
  {
  \begin{minipage}{.23\textwidth}
    \centering
    {\includegraphics[width=\linewidth]{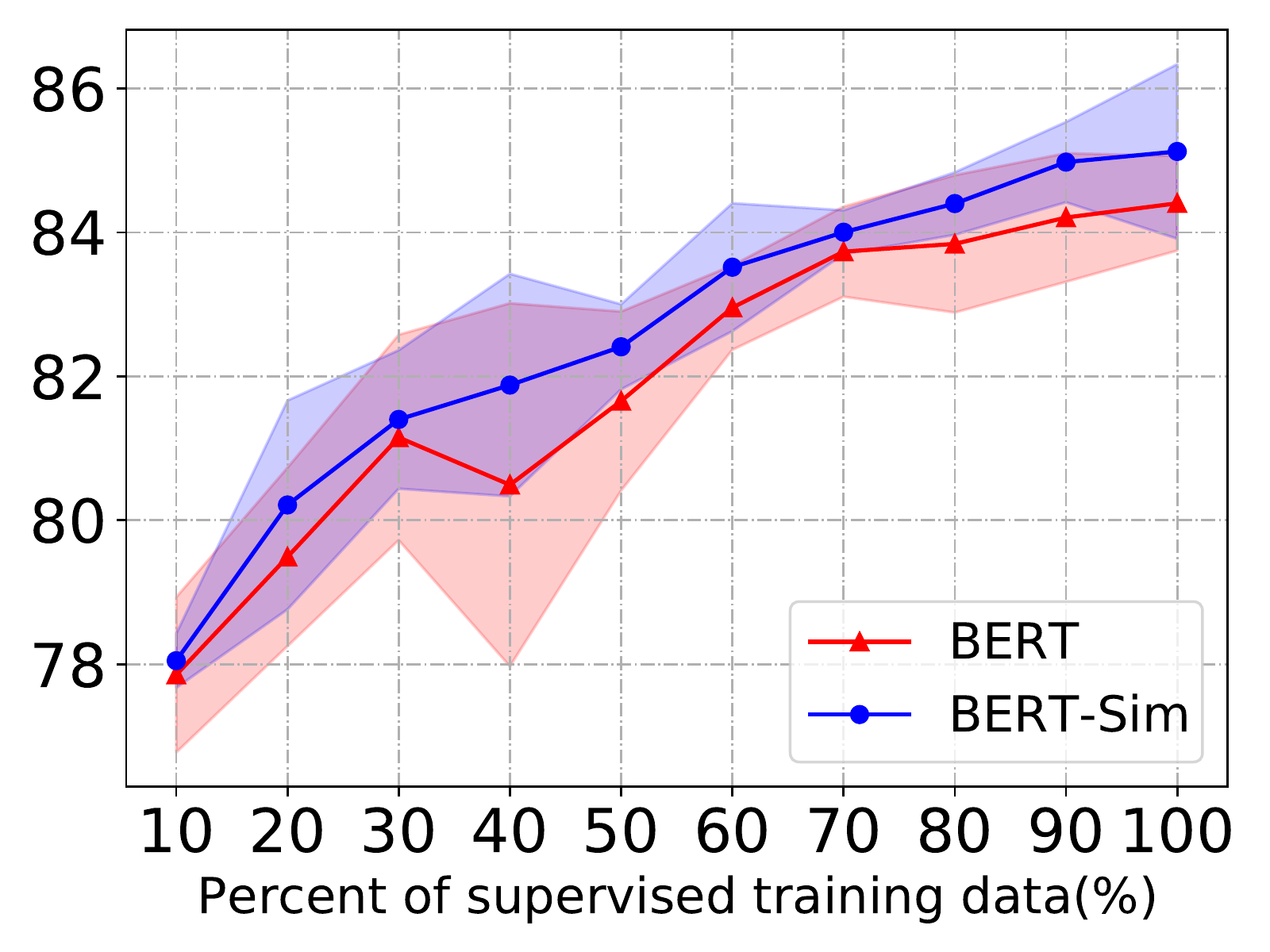}}
    \label{fig:illustrate_cases}
  \end{minipage}
  }
  \subfigure[QQP]
  {
  \begin{minipage}{.23\textwidth}
    \centering
    {\includegraphics[width=\linewidth]{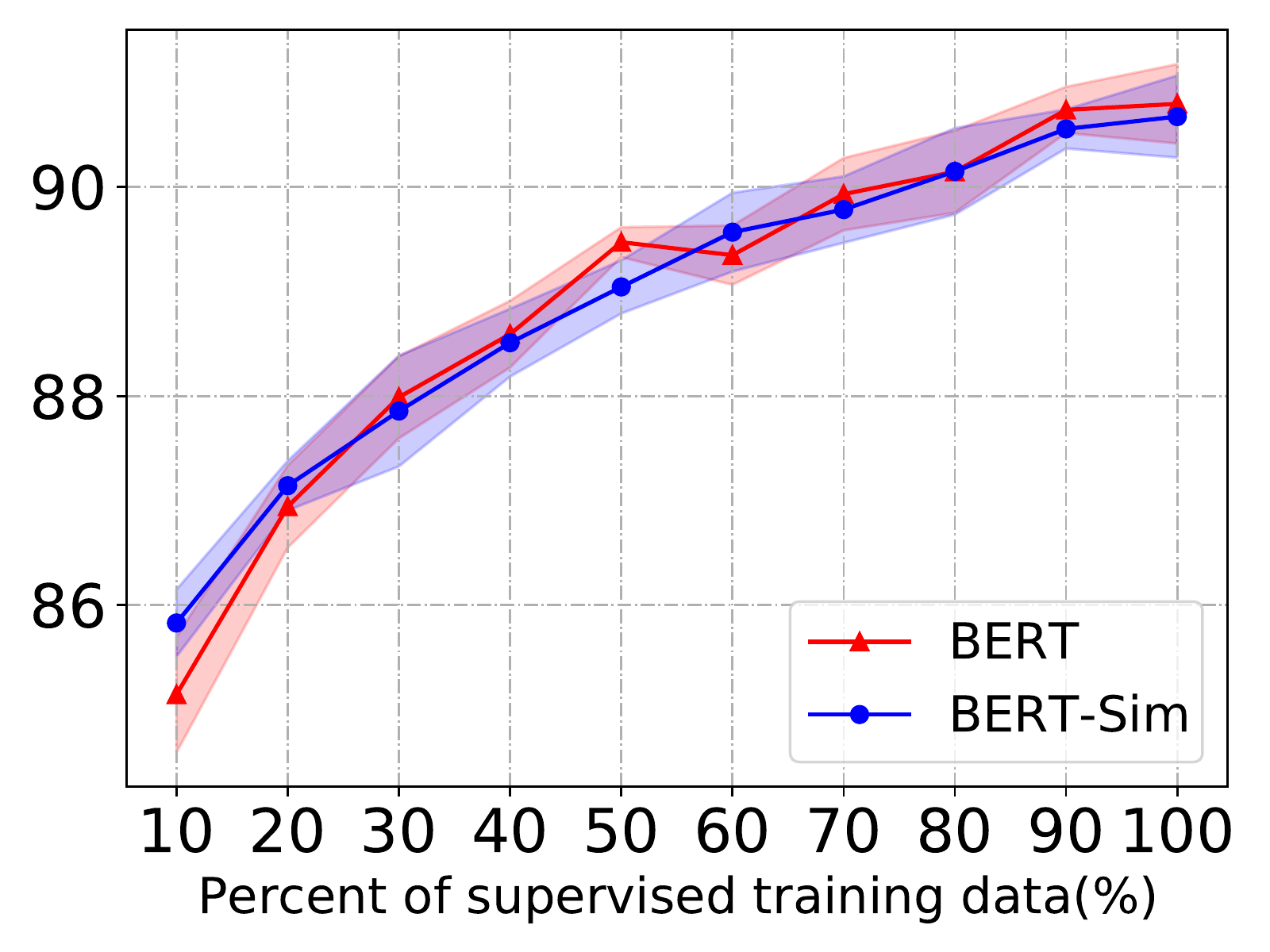}}
    \label{fig:illustrate_cases}
  \end{minipage}
  }
  \subfigure[TwitterURL]
  {
  \begin{minipage}{.23\textwidth}
    \centering
    {\includegraphics[width=\linewidth]{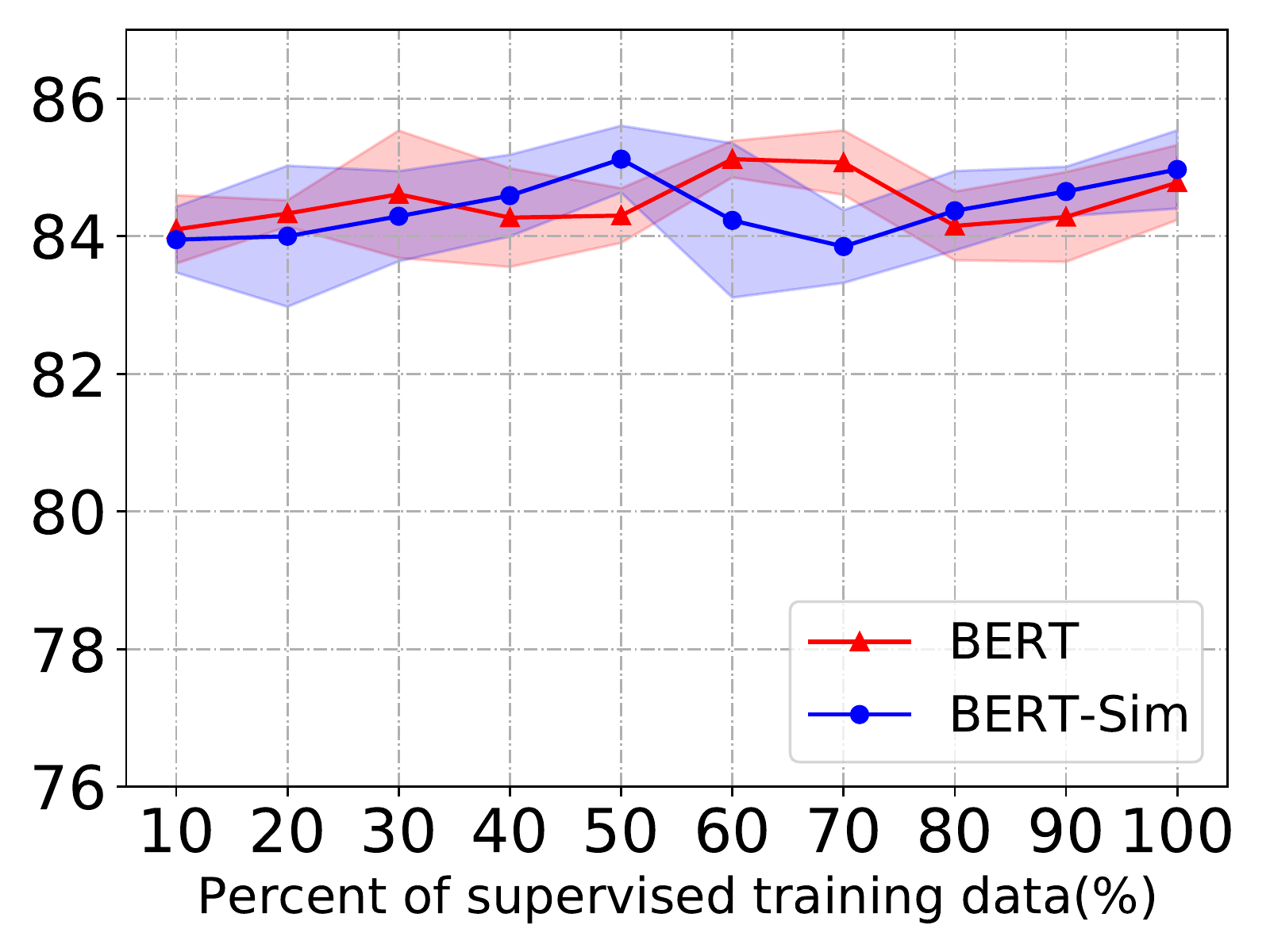}}
    \label{fig:voronoi}
  \end{minipage}
  }
  \caption{Performance of BERT and BERT-Sim with different amounts of training data. X-axis: Percent of supervised training data. Y-axis: Macro-F1 for MRPC and Twitter-URL, Accuracy for QQP and Pearson Correlation for STS-B . The colored bands indicate $\pm 1$ standard deviation corresponding to different percentages of training data.}
  \end{figure*}

 \begin{figure*}[htbp]
  \centering
  \subfigure[MRPC]
  {
  \begin{minipage}{.23\textwidth}
    \centering
    {\includegraphics[width=\linewidth]{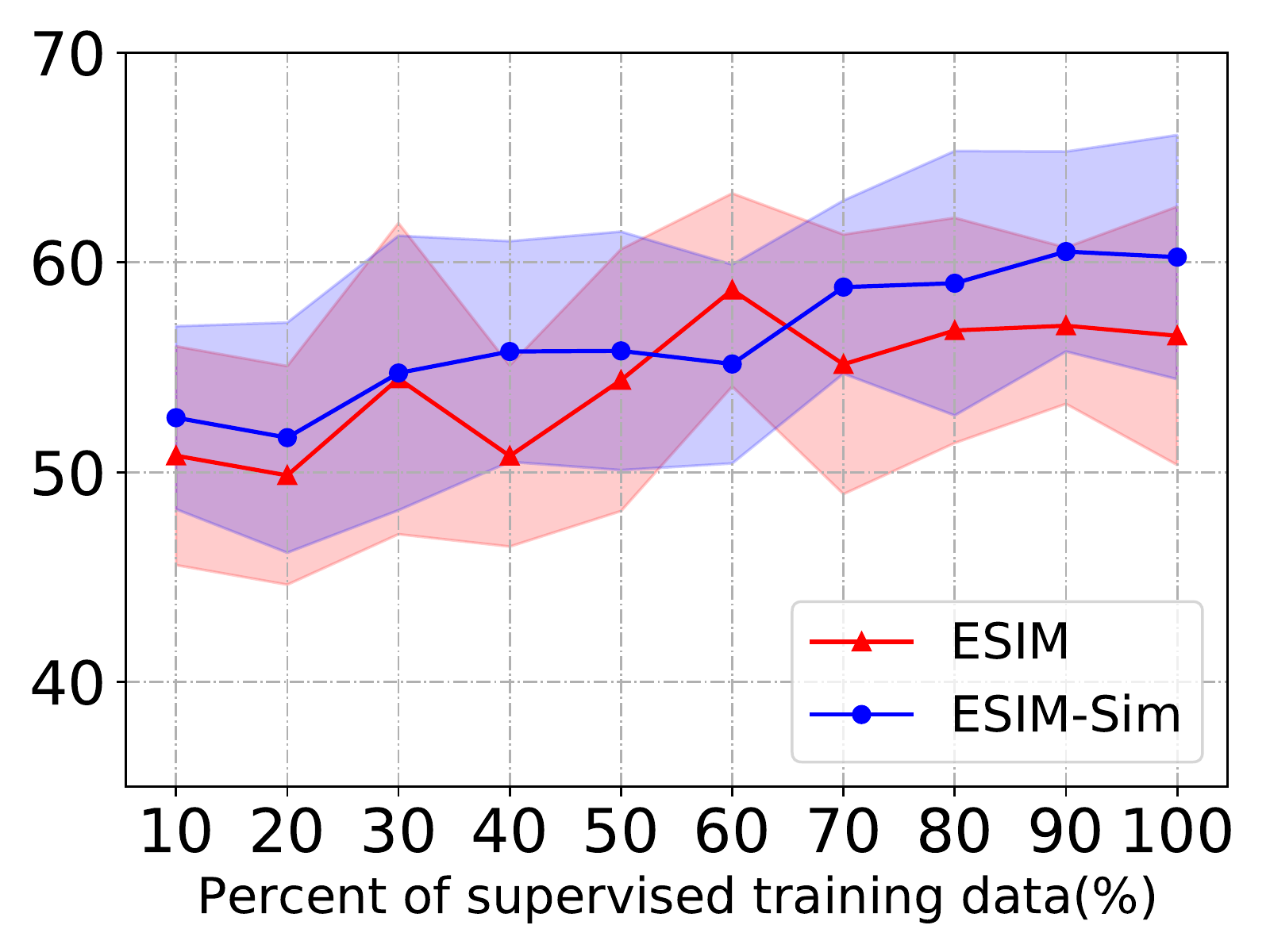}}
    \label{fig:illustrate_cases}
  \end{minipage}
  }
  \subfigure[STS-B]
  {
  \begin{minipage}{.23\textwidth}
    \centering
    {\includegraphics[width=\linewidth]{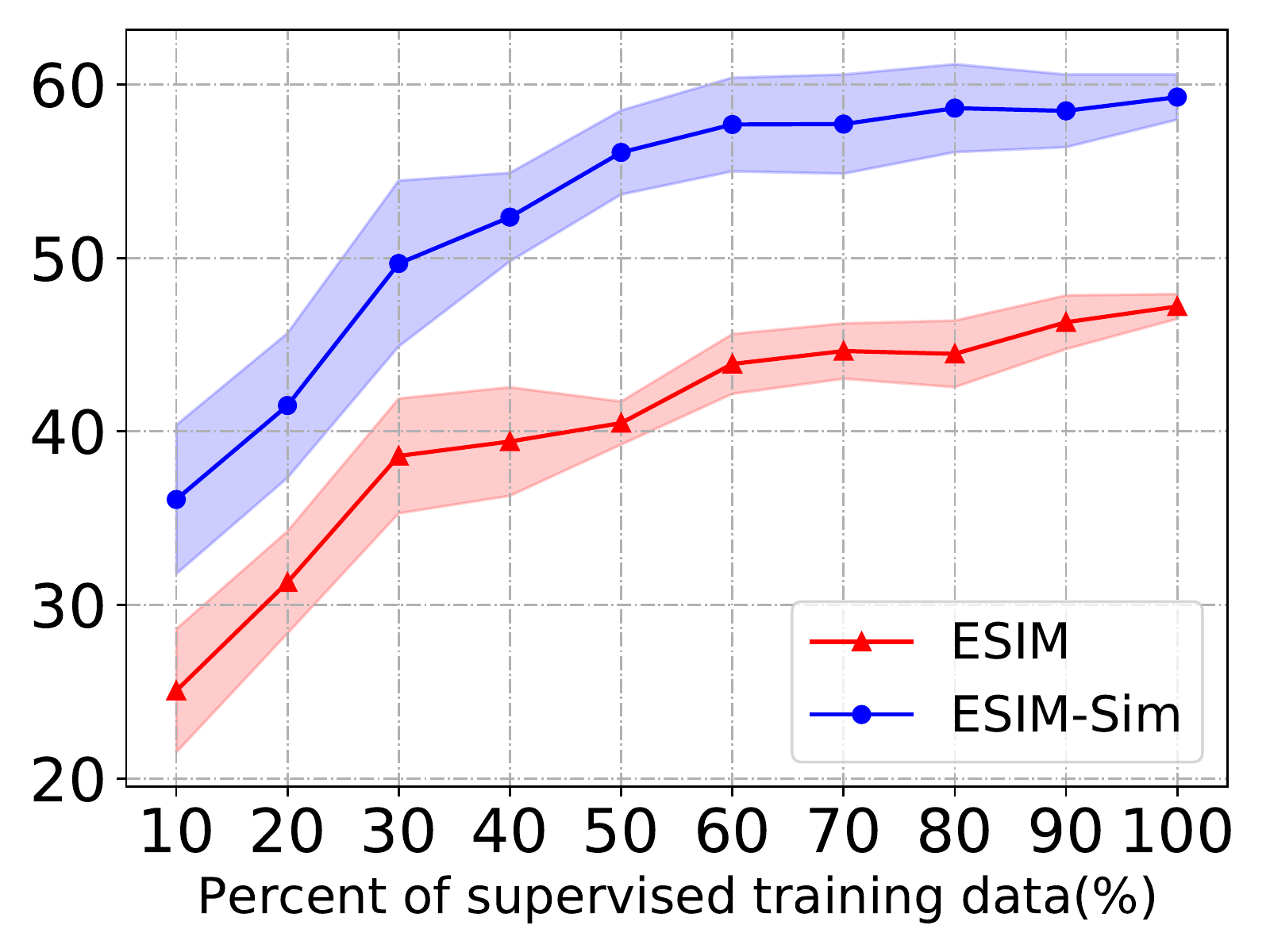}}
    \label{fig:illustrate_cases}
  \end{minipage}
  }
  \subfigure[QQP]
  {
  \begin{minipage}{.23\textwidth}
    \centering
    {\includegraphics[width=\linewidth]{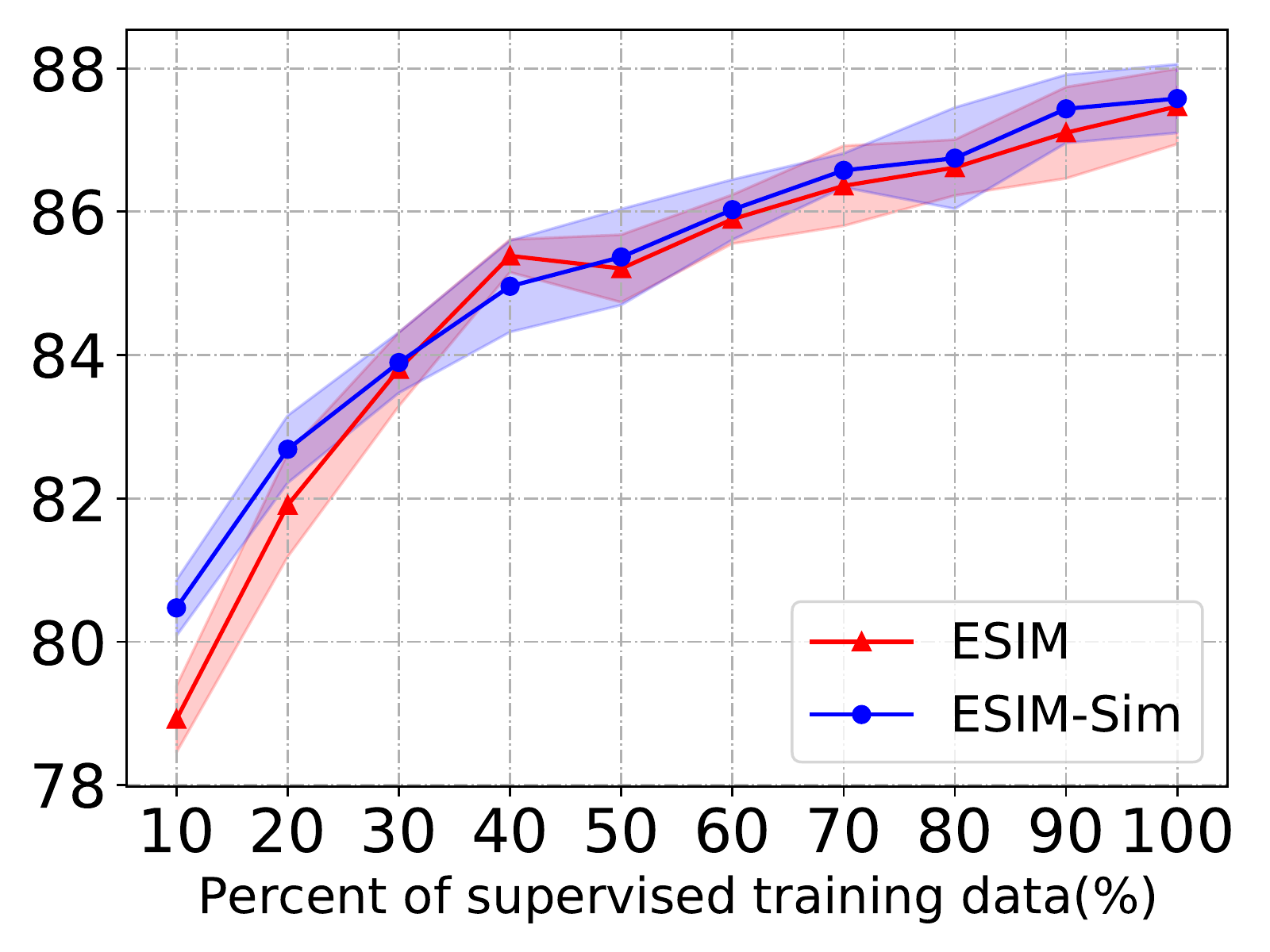}}
    \label{fig:illustrate_cases}
  \end{minipage}
  }
  \subfigure[TwitterURL]
  {
  \begin{minipage}{.23\textwidth}
    \centering
    {\includegraphics[width=\linewidth]{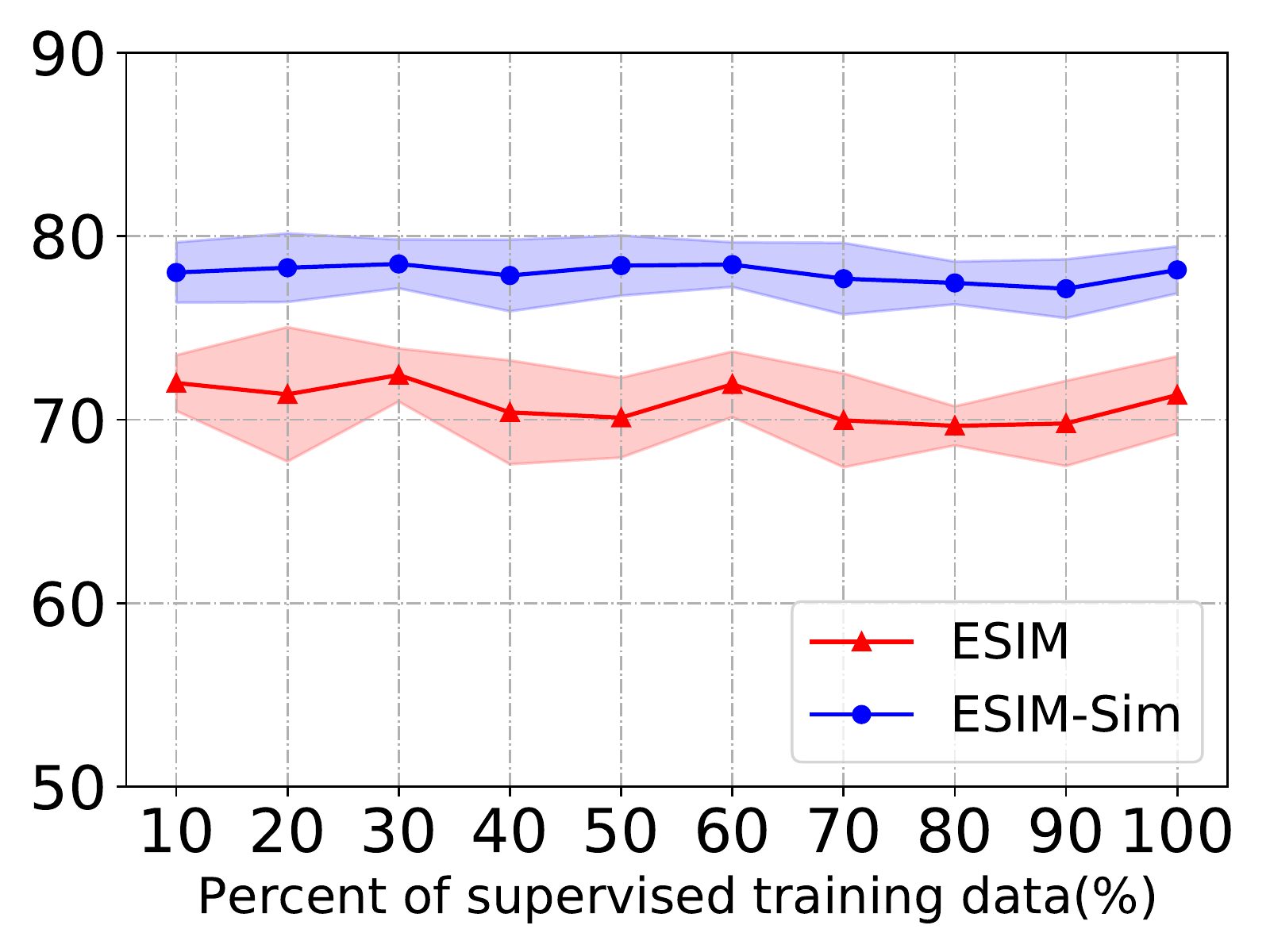}}
    \label{fig:voronoi}
  \end{minipage}
  }
  \caption{Performance of ESIM and ESIM-Sim with different amounts of training data. X-axis: Percent of supervised training data. Y-axis: Macro-F1 for MRPC and Twitter-URL, Accuracy for QQP and Pearson Correlation for STS-B. The colored bands indicate $\pm 1$ standard deviation corresponding to different percentages of training data.}
  \end{figure*} 

  \begin{figure}
  \centering
   \includegraphics[width=.65\linewidth]{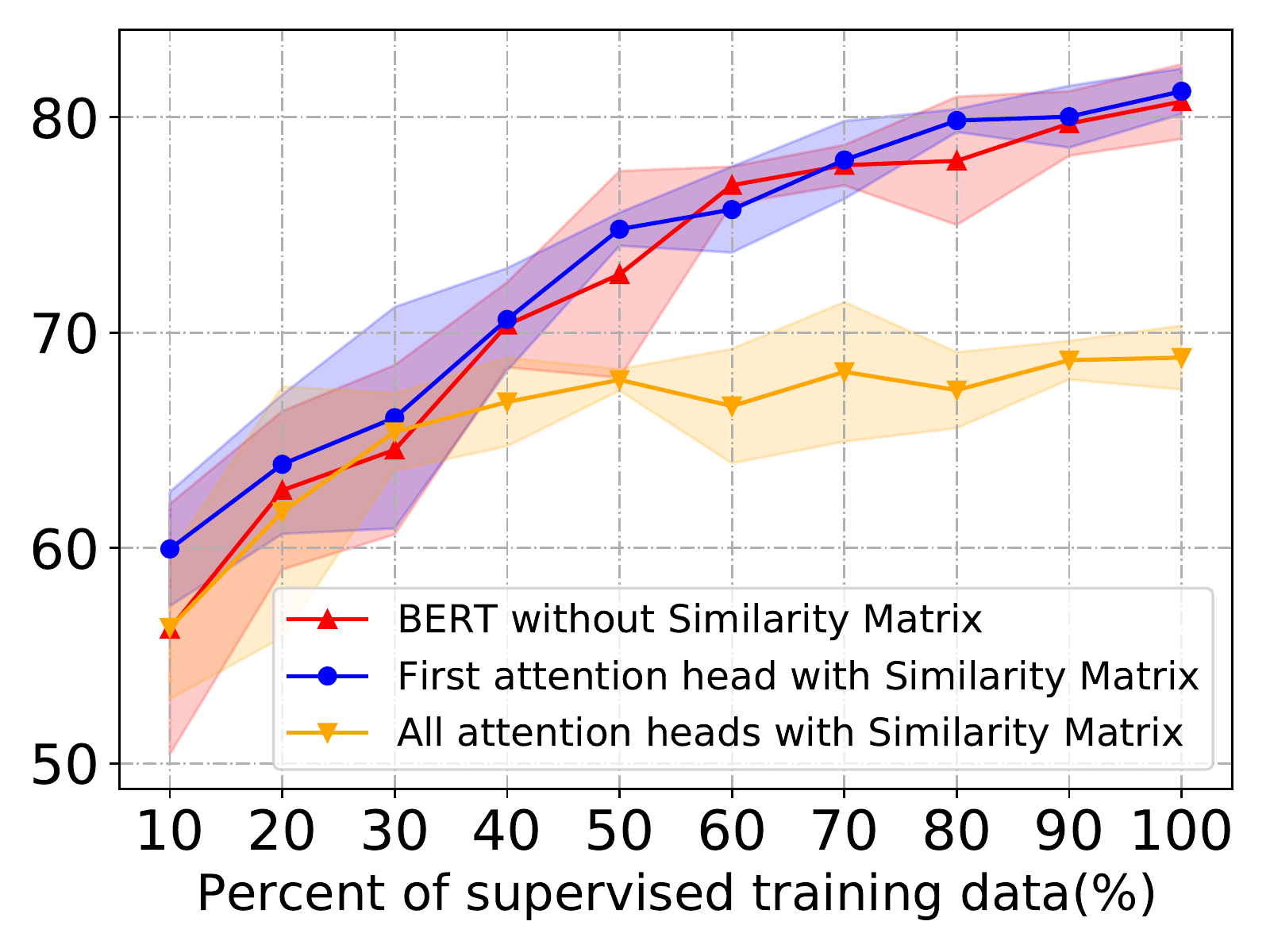}
  \caption{The effect of adding prior knowledge to different attention heads on the model performance on MRPC dataset. The y-axis represents Macro-F1.}
  \label{fig:teaser}
  \end{figure}

We add prior knowledge to both ESIM model and BERT model and train them on the above four datasets. 
Table \ref{tb:benchmark_stat} shows the results with our implementation. We use accuracy, F$_1$ score of the positive class, macro-F$_1$ score, Pearson correlation, Spearman correlation for evaluation on different datasets. These  metrics follow previous literature \cite{wang2017bilateral, devlin2018bert, wang2019glue}. Compared with the BERT model, ESIM model has a faster training speed and takes up less memory, so in Table \ref{tb:benchmark_stat} the results of ESIM model and ESIM-Sim model are the average after training 10 times. For BERT model, due to its slow training speed and large memory usage, the results are the average over 5 runs. 
In addition, our results have slight deviation from the results reported in the Google BERT paper \cite{devlin2018bert}. We suspect the following potential reasons: (i) The training set and the validation set are split in different ways, for example QQP dataset. (ii) The random shuffling of the training set causes the deviation of the test result.

\subsection{Learning Curve Analysis}
At the same time, for each dataset, we randomly select from 10\% to 100\% data from training set as training data. For BERT and BERT-Sim, we have trained 5 times for each training scale of each dataset. We show the results in Figure 7. Similarly, for ESIM and ESIM-Sim, we have trained 10 runs; the results are shown in Figure 8.

To verify our hypothesis that word similarity knowledge is most needed in the first layer, we explored an alternative approach for comparison -- to inject this knowledge into BERT's attention on \emph{all} layers. We used the MRPC dataset in this analysis. The learning curve is shown in Figure 9.

\subsection{Discussion}
\begin{table*}[t!]
\centering
\caption{Example sentence pairs in the TwitterURL dataset. The four sentence pairs shares a common sentence.}
\begin{tabular}{@{}ll@{}}
\toprule
\textbf{Sentence pair} &
  \textbf{Label} \\ \midrule
\begin{tabular}[c]{@{}l@{}} $s_{a}$: Scientists create 3D-printed objects that can change shape after they are printed.\\ $s_{b}$: Shape-shifting 3D printed objects are now a thing.\end{tabular} &
  Paraphrase \\ \midrule
\begin{tabular}[c]{@{}l@{}} $s_{a}$: Scientists create 3D-printed objects that can change shape after they are printed.\\ $s_{b}$: 3D-printed objects change shape after printed an open door to new use cases.\end{tabular} &
  Paraphrase \\ \midrule
\begin{tabular}[c]{@{}l@{}} $s_{a}$: Scientists create 3D-printed objects that can change shape after they are printed.\\ $s_{b}$: Imagine a drunk guy in the bar's bathroom , watching the urinal cakes morph.\end{tabular} &
  Non-paraphrase \\ \midrule
\begin{tabular}[c]{@{}l@{}} $s_{a}$: Scientists create 3D-printed objects that can change shape after they are printed.\\ $s_{b}$: 4D printed materials are the future of architecture . So many great potential applications , esp. in green design.\end{tabular} &
  Non-paraphrase \\ \bottomrule
\end{tabular}
\label{tab:twitter-url-examples}
\end{table*}

In Table 1, on the two relatively small datasets (MRPC and STS-B), we observe salient performance gain when knowledge is incorporated in the model (ESIM vs. ESIM-Sim; BERT vs. BERT-Sim). On the two relatively large datasets (QQP and TwitterURL), the performance gain is minimal if any. This is what we expected -- prior knowledge is most beneficial when the training data is small. Comparing the performance gain within each model family, we observe that  BERT gains less performance than ESIM when the word similarity knowledge is added. This echos with our finding in Section \ref{sec:bertanalysis} that a pre-trained BERT already contains much knowledge about  semantic textual matching tasks. In comparison, ESIM lacks the same level of knowledge even it is initialized with pre-trained word vectors.

Across the datasets, we observe that the gain in F$_1$ score is mainly caused by a salient increase in precision and a negligible decrease in recall. In other words, prior knowledge mainly helped BERT reduce false positives on the semantic text similarity task. This implies that the original BERT may not be able to tell the subtle semantic difference between a pair of related (but not synonymous) words, and therefore wrongly classified non-paraphrase sentences as paraphrases.

The learning curves in Figure 7 and 8 reveal a number of interesting patterns. Our proposed method for adding prior knowledge improves model performance almost consistently across all training data sizes. It is encouraging to see that the injected prior knowledge did not impose an unnecessarily strong inductive bias (in the sense of bias-variance trade-off) to the models. In other words, the prior knowledge is consistent with the learning objective and beneficial to the models all along the learning process.

On  MRPC, STS-B, and QQP datasets, prior knowledge provides the most salient performance gain happens when the training data is small, and the ESIM gains more than BERT. This is in agreement with our observation in Table 1. Also note that BERT trained on 10\% to 20\% of the training data can already outperform ESIM trained on full training data, and BERT gains performance at a faster rate than ESIM. This again highlights the  rich knowledge and superior learning capability of BERT, compared to non-Transformer models. These findings suggest that it is most sensible to consider adding knowledge to BERT if the training data is scarce, i.e. on the order of hundreds of sentence pairs for a STS task.

In Figure 9, we compared the results between adding prior knowledge to the first layer of BERT's multi-head attention vs. adding prior knowledge to all layers, on the MRPC dataset. We observe that adding knowledge to all layers will impede the model from learning, creating an undesirable inductive bias. In contrast, adding knowledge to the first layer is more reasonable. This indicates that the broad/uniform attention on lower layers of BERT needs to be guided, while in higher layers attention heads have dedicated roles in collecting specific syntactic and semantic information.

What's peculiar is that on the TwitterURL dataset, we do not see similar results as the other datasets. We speculate this is mainly related to the form of the dataset. We show part of the TwitterURL dataset in detail in Table \ref{tab:twitter-url-examples}. In the TwitterURL dataset, it often happens that the same sentence $s_{a}$ corresponds to many different $s_{b}$'s, so when we increase the training data from 10\% to 100\%, a model does not see many completely new sentence pairs -- it may have already seen at least one of the two sentences.
This explains why the results obtained when the training data is 10\% of the entire training set are similar to the results on 100\% training data.

In Section 3, we use synonym knowledge to augment  the MRPC dataset and train a BERT model that achieves 88.0 $\pm$ 0.4 \% F1 score. In Table 1, BERT-Sim achieves 88.3 $\pm$ 0.3 \% F1 score using the same form of knowledge. This suggests BERT-Sim performs at the same level as (or slightly better than) a data-augmented BERT.
In Table \ref{tab:aug-vs-sim}, we show this type of comparison on all four datasets. BERT-sim slightly outperform data augmentation because it can inject word similarity knowledge -- e.g. ``happy'' and ``glad'' are synonyms -- into the model even if those words do not appear in the training data. Whereas for data augmentation, if no sentence in the training data contains the word ``happy'' or ``glad'', then the knowledge that ``happy'' and ``glad'' are synonyms cannot be added by  data augmentation, i.e., replacing synonyms in training data. Therefore the benefit of data augmentation depends on training data, while the benefit of knowledge injection does not. 

\begin{table}[h]
\centering
\caption{Performance in four datasets (\%) using data augmentation and knowledge injection. We report for one metric per dataset as the other metric has similar trends. Numbers in parentheses are standard deviations.}
\begin{tabular}{l|cc}
\toprule
 \textbf{Dataset / metric}         & \textbf{Data-augmented BERT}  & \textbf{BERT-Sim} \\ \hline
MRPC / F1 &88.0 (0.4) &88.3 (0.3)\\ 
STS-B / Pearson corr. &84.4 (0.7) &85.1 (1.2)\\ 
QQP / Accuracy  &90.6 (0.3) &90.7 (0.4)\\ 
TwitterURL / F1 &75.9 (0.7) &76.2 (0.7)\\ \bottomrule
\end{tabular}
\label{tab:aug-vs-sim}
\end{table}

Note that because of extra training data, the data augmentation approach takes twice as much time to train than directly injecting knowledge into the model.  Table \ref{tab:train-time} shows per-epoch training time on four datasets. This highlights an advantage of our proposed approach -- it performs at least as well as the data augmentation approach with only half the training time.

\begin{table}[h]
\centering
\caption{Per-epoch training time  (in seconds) on the four datasets using data augmentation and knowledge injection.}
\begin{tabular}{l|cc}
\toprule
 \textbf{Dataset}         & \textbf{Data-augmented BERT}  & \textbf{BERT-Sim} \\ \hline
MRPC &73 &36\\ 
STS-B &99 &53\\ 
QQP &6372 &3206\\ 
TwitterURL &775 &338\\ \bottomrule
\end{tabular}
\label{tab:train-time}
\end{table}
\section{Conclusion}
In this paper, by analyzing what BERT has already known, what task-specific knowledge BERT needs and where it needs it, we proposed an effective and efficient method for injecting word similarity knowledge into BERT without adding new training task but directly guiding model's attention. This method is also applicable to non-Transformer deep model. Through experiments on different scale datasets on the semantic textual similarity (STS) task, we prove that the prior knowledge of word similarity is able to consistently improve the STS performance of BERT and the benefit is especially salient when training data is scarce. 

\vspace{+.2in}
\section*{Acknowledgement}
The authors would like to thank the anonymous referees for their valuable comments. This work is supported by the  National Natural Science Foundation of China (No. 61976102 and No. U19A2065) and the Fundamental Research Funds for the Central Universities, JLU.

\bibliographystyle{ACM-Reference-Format}
\bibliography{ref}


\begin{thebibliography}{48}


\ifx \showCODEN    \undefined \def \showCODEN     #1{\unskip}     \fi
\ifx \showDOI      \undefined \def \showDOI       #1{#1}\fi
\ifx \showISBNx    \undefined \def \showISBNx     #1{\unskip}     \fi
\ifx \showISBNxiii \undefined \def \showISBNxiii  #1{\unskip}     \fi
\ifx \showISSN     \undefined \def \showISSN      #1{\unskip}     \fi
\ifx \showLCCN     \undefined \def \showLCCN      #1{\unskip}     \fi
\ifx \shownote     \undefined \def \shownote      #1{#1}          \fi
\ifx \showarticletitle \undefined \def \showarticletitle #1{#1}   \fi
\ifx \showURL      \undefined \def \showURL       {\relax}        \fi
\providecommand\bibfield[2]{#2}
\providecommand\bibinfo[2]{#2}
\providecommand\natexlab[1]{#1}
\providecommand\showeprint[2][]{arXiv:#2}

\bibitem[\protect\citeauthoryear{Beltagy, Roller, Cheng, Erk, and
  Mooney}{Beltagy et~al\mbox{.}}{2015}]%
        {beltagy2015representing}
\bibfield{author}{\bibinfo{person}{Islam Beltagy}, \bibinfo{person}{Stephen
  Roller}, \bibinfo{person}{Pengxiang Cheng}, \bibinfo{person}{Katrin Erk},
  {and} \bibinfo{person}{Raymond~J. Mooney}.} \bibinfo{year}{2015}\natexlab{}.
\newblock \showarticletitle{Representing Meaning with a Combination of Logical
  Form and Vectors}.
\newblock \bibinfo{journal}{\emph{CoRR}}  \bibinfo{volume}{abs/1505.06816}
  (\bibinfo{year}{2015}).
\newblock


\bibitem[\protect\citeauthoryear{Bodenreider}{Bodenreider}{2004}]%
        {bodenreider2004unified}
\bibfield{author}{\bibinfo{person}{Olivier Bodenreider}.}
  \bibinfo{year}{2004}\natexlab{}.
\newblock \showarticletitle{The unified medical language system (UMLS):
  integrating biomedical terminology}.
\newblock \bibinfo{journal}{\emph{Nucleic acids research}}
  \bibinfo{volume}{32}, \bibinfo{number}{suppl\_1} (\bibinfo{year}{2004}),
  \bibinfo{pages}{D267--D270}.
\newblock


\bibitem[\protect\citeauthoryear{Cer, Diab, Agirre, Lopez-Gazpio, and
  Specia}{Cer et~al\mbox{.}}{2017}]%
        {cer2017semeval}
\bibfield{author}{\bibinfo{person}{Daniel Cer}, \bibinfo{person}{Mona Diab},
  \bibinfo{person}{Eneko Agirre}, \bibinfo{person}{Inigo Lopez-Gazpio}, {and}
  \bibinfo{person}{Lucia Specia}.} \bibinfo{year}{2017}\natexlab{}.
\newblock \showarticletitle{Semeval-2017 task 1: Semantic textual
  similarity-multilingual and cross-lingual focused evaluation}.
\newblock \bibinfo{journal}{\emph{arXiv preprint arXiv:1708.00055}}
  (\bibinfo{year}{2017}).
\newblock


\bibitem[\protect\citeauthoryear{Chen, Hu, Liu, Xiao, and Jiang}{Chen
  et~al\mbox{.}}{2019}]%
        {chen2019deep}
\bibfield{author}{\bibinfo{person}{Jindong Chen}, \bibinfo{person}{Yizhou Hu},
  \bibinfo{person}{Jingping Liu}, \bibinfo{person}{Yanghua Xiao}, {and}
  \bibinfo{person}{Haiyun Jiang}.} \bibinfo{year}{2019}\natexlab{}.
\newblock \showarticletitle{Deep short text classification with knowledge
  powered attention}. In \bibinfo{booktitle}{\emph{Proceedings of the AAAI
  Conference on Artificial Intelligence}}, Vol.~\bibinfo{volume}{33}.
  \bibinfo{pages}{6252--6259}.
\newblock


\bibitem[\protect\citeauthoryear{Chen, Zhu, Ling, Wei, Jiang, and Inkpen}{Chen
  et~al\mbox{.}}{2017}]%
        {chen2017enhanced}
\bibfield{author}{\bibinfo{person}{Qian Chen}, \bibinfo{person}{Xiaodan Zhu},
  \bibinfo{person}{Zhenhua Ling}, \bibinfo{person}{Si Wei},
  \bibinfo{person}{Hui Jiang}, {and} \bibinfo{person}{Diana Inkpen}.}
  \bibinfo{year}{2017}\natexlab{}.
\newblock \showarticletitle{Enhanced LSTM for Natural Language Inference}. In
  \bibinfo{booktitle}{\emph{Proceedings of the 55th Annual Meeting of the
  Association for Computational Linguistics (ACL 2017)}}.
  \bibinfo{publisher}{ACL}, \bibinfo{address}{Vancouver},
  \bibinfo{pages}{1657--1668}.
\newblock


\bibitem[\protect\citeauthoryear{Chen, Zhu, Ling, Inkpen, and Wei}{Chen
  et~al\mbox{.}}{2018}]%
        {chen2018neural}
\bibfield{author}{\bibinfo{person}{Qian Chen}, \bibinfo{person}{Xiaodan Zhu},
  \bibinfo{person}{Zhen-Hua Ling}, \bibinfo{person}{Diana Inkpen}, {and}
  \bibinfo{person}{Si Wei}.} \bibinfo{year}{2018}\natexlab{}.
\newblock \showarticletitle{Neural Natural Language Inference Models Enhanced
  with External Knowledge}. In \bibinfo{booktitle}{\emph{Proceedings of the
  56th Annual Meeting of the Association for Computational Linguistics (Volume
  1: Long Papers)}}. \bibinfo{publisher}{Association for Computational
  Linguistics}, \bibinfo{pages}{2406--2417}.
\newblock


\bibitem[\protect\citeauthoryear{Chen, Hakkani-Tur, Tur, Celikyilmaz, Gao, and
  Deng}{Chen et~al\mbox{.}}{2016}]%
        {chen2016knowledge}
\bibfield{author}{\bibinfo{person}{Yun-Nung Chen}, \bibinfo{person}{Dilek
  Hakkani-Tur}, \bibinfo{person}{Gokhan Tur}, \bibinfo{person}{Asli
  Celikyilmaz}, \bibinfo{person}{Jianfeng Gao}, {and} \bibinfo{person}{Li
  Deng}.} \bibinfo{year}{2016}\natexlab{}.
\newblock \showarticletitle{Knowledge as a teacher: Knowledge-guided structural
  attention networks}.
\newblock \bibinfo{journal}{\emph{arXiv preprint arXiv:1609.03286}}
  (\bibinfo{year}{2016}).
\newblock


\bibitem[\protect\citeauthoryear{Clark, Khandelwal, Levy, and Manning}{Clark
  et~al\mbox{.}}{2019}]%
        {clark2019what}
\bibfield{author}{\bibinfo{person}{Kevin Clark}, \bibinfo{person}{Urvashi
  Khandelwal}, \bibinfo{person}{Omer Levy}, {and}
  \bibinfo{person}{Christopher~D. Manning}.} \bibinfo{year}{2019}\natexlab{}.
\newblock \showarticletitle{What Does BERT Look At? An Analysis of BERT's
  Attention}. In \bibinfo{booktitle}{\emph{BlackBoxNLP@ACL}}.
\newblock


\bibitem[\protect\citeauthoryear{Das and Smith}{Das and Smith}{2009}]%
        {das2009paraphrase}
\bibfield{author}{\bibinfo{person}{Dipanjan Das} {and} \bibinfo{person}{Noah~A
  Smith}.} \bibinfo{year}{2009}\natexlab{}.
\newblock \showarticletitle{Paraphrase identification as probabilistic
  quasi-synchronous recognition}. In \bibinfo{booktitle}{\emph{Proceedings of
  the Joint Conference of the 47th Annual Meeting of the ACL and the 4th
  International Joint Conference on Natural Language Processing of the AFNLP}}.
  \bibinfo{pages}{468--476}.
\newblock


\bibitem[\protect\citeauthoryear{Devlin, Chang, Lee, and Toutanova}{Devlin
  et~al\mbox{.}}{2018}]%
        {devlin2018bert}
\bibfield{author}{\bibinfo{person}{Jacob Devlin}, \bibinfo{person}{Ming-Wei
  Chang}, \bibinfo{person}{Kenton Lee}, {and} \bibinfo{person}{Kristina
  Toutanova}.} \bibinfo{year}{2018}\natexlab{}.
\newblock \showarticletitle{BERT: Pre-training of Deep Bidirectional
  Transformers for Language Understanding}.
\newblock \bibinfo{journal}{\emph{arXiv preprint arXiv:1810.04805}}
  (\bibinfo{year}{2018}).
\newblock


\bibitem[\protect\citeauthoryear{Dolan and Brockett}{Dolan and
  Brockett}{2005}]%
        {dolan2005automatically}
\bibfield{author}{\bibinfo{person}{William~B Dolan} {and}
  \bibinfo{person}{Chris Brockett}.} \bibinfo{year}{2005}\natexlab{}.
\newblock \showarticletitle{Automatically Constructing a Corpus of Sentential
  Paraphrases}. In \bibinfo{booktitle}{\emph{Proceedings of the Third
  International Workshop on Paraphrasing (IWP2005)}}.
\newblock


\bibitem[\protect\citeauthoryear{Ettinger}{Ettinger}{2020}]%
        {ettinger2020bert}
\bibfield{author}{\bibinfo{person}{Allyson Ettinger}.}
  \bibinfo{year}{2020}\natexlab{}.
\newblock \showarticletitle{What BERT is not: Lessons from a new suite of
  psycholinguistic diagnostics for language models}.
\newblock \bibinfo{journal}{\emph{Transactions of the Association for
  Computational Linguistics}}  \bibinfo{volume}{8} (\bibinfo{year}{2020}),
  \bibinfo{pages}{34--48}.
\newblock


\bibitem[\protect\citeauthoryear{Fernando and Stevenson}{Fernando and
  Stevenson}{2008}]%
        {fernando2008semantic}
\bibfield{author}{\bibinfo{person}{Samuel Fernando} {and} \bibinfo{person}{Mark
  Stevenson}.} \bibinfo{year}{2008}\natexlab{}.
\newblock \showarticletitle{A semantic similarity approach to paraphrase
  detection}. In \bibinfo{booktitle}{\emph{Proceedings of the 11th annual
  research colloquium of the UK special interest group for computational
  linguistics}}. \bibinfo{pages}{45--52}.
\newblock


\bibitem[\protect\citeauthoryear{Gururangan, Marasovi{\'c}, Swayamdipta, Lo,
  Beltagy, Downey, and Smith}{Gururangan et~al\mbox{.}}{2020}]%
        {gururangan2020dont}
\bibfield{author}{\bibinfo{person}{Suchin Gururangan}, \bibinfo{person}{Ana
  Marasovi{\'c}}, \bibinfo{person}{Swabha Swayamdipta}, \bibinfo{person}{Kyle
  Lo}, \bibinfo{person}{Iz Beltagy}, \bibinfo{person}{Doug Downey}, {and}
  \bibinfo{person}{Noah~A. Smith}.} \bibinfo{year}{2020}\natexlab{}.
\newblock \showarticletitle{Don{'}t Stop Pretraining: Adapt Language Models to
  Domains and Tasks}. In \bibinfo{booktitle}{\emph{Proceedings of the 58th
  Annual Meeting of the Association for Computational Linguistics}}.
  \bibinfo{publisher}{Association for Computational Linguistics},
  \bibinfo{address}{Online}, \bibinfo{pages}{8342--8360}.
\newblock


\bibitem[\protect\citeauthoryear{He, Liu, Gao, and Chen}{He
  et~al\mbox{.}}{2020}]%
        {he2020deberta}
\bibfield{author}{\bibinfo{person}{Pengcheng He}, \bibinfo{person}{Xiaodong
  Liu}, \bibinfo{person}{Jianfeng Gao}, {and} \bibinfo{person}{Weizhu Chen}.}
  \bibinfo{year}{2020}\natexlab{}.
\newblock \bibinfo{title}{DeBERTa: Decoding-enhanced BERT with Disentangled
  Attention}.
\newblock
\newblock
\showeprint[arxiv]{cs.CL/2006.03654}


\bibitem[\protect\citeauthoryear{Hewitt and Manning}{Hewitt and
  Manning}{2019}]%
        {hewitt2019structural}
\bibfield{author}{\bibinfo{person}{John Hewitt} {and}
  \bibinfo{person}{Christopher~D Manning}.} \bibinfo{year}{2019}\natexlab{}.
\newblock \showarticletitle{A structural probe for finding syntax in word
  representations}. In \bibinfo{booktitle}{\emph{Proceedings of the 2019
  Conference of the North American Chapter of the Association for Computational
  Linguistics: Human Language Technologies, Volume 1 (Long and Short Papers)}}.
  \bibinfo{pages}{4129--4138}.
\newblock


\bibitem[\protect\citeauthoryear{Hochreiter and Schmidhuber}{Hochreiter and
  Schmidhuber}{1997}]%
        {hochreiter1997long}
\bibfield{author}{\bibinfo{person}{Sepp Hochreiter} {and}
  \bibinfo{person}{J{\"u}rgen Schmidhuber}.} \bibinfo{year}{1997}\natexlab{}.
\newblock \showarticletitle{Long short-term memory}.
\newblock \bibinfo{journal}{\emph{Neural computation}} \bibinfo{volume}{9},
  \bibinfo{number}{8} (\bibinfo{year}{1997}), \bibinfo{pages}{1735--1780}.
\newblock


\bibitem[\protect\citeauthoryear{Hu, Lu, Li, and Chen}{Hu
  et~al\mbox{.}}{2014}]%
        {hu2014convolutional}
\bibfield{author}{\bibinfo{person}{Baotian Hu}, \bibinfo{person}{Zhengdong Lu},
  \bibinfo{person}{Hang Li}, {and} \bibinfo{person}{Qingcai Chen}.}
  \bibinfo{year}{2014}\natexlab{}.
\newblock \showarticletitle{Convolutional neural network architectures for
  matching natural language sentences}. In \bibinfo{booktitle}{\emph{Advances
  in neural information processing systems}}. \bibinfo{pages}{2042--2050}.
\newblock


\bibitem[\protect\citeauthoryear{Iftene and Balahur}{Iftene and
  Balahur}{2007}]%
        {iftene2007hypothesis}
\bibfield{author}{\bibinfo{person}{Adrian Iftene} {and}
  \bibinfo{person}{Alexandra Balahur}.} \bibinfo{year}{2007}\natexlab{}.
\newblock \showarticletitle{Hypothesis transformation and semantic variability
  rules used in recognizing textual entailment}. In
  \bibinfo{booktitle}{\emph{Proceedings of the ACL-PASCAL Workshop on Textual
  Entailment and Paraphrasing}}. \bibinfo{pages}{125--130}.
\newblock


\bibitem[\protect\citeauthoryear{Iyer, Dandekar, and Csernai}{Iyer
  et~al\mbox{.}}{2017}]%
        {iyer2017first}
\bibfield{author}{\bibinfo{person}{Shankar Iyer}, \bibinfo{person}{Nikhil
  Dandekar}, {and} \bibinfo{person}{Korn{\'e}l Csernai}.}
  \bibinfo{year}{2017}\natexlab{}.
\newblock \showarticletitle{First quora dataset release: Question pairs}.
\newblock \bibinfo{journal}{\emph{URL
  https://data.quora.com/First-Quora-Dataset-Release-Question-Pairs}}
  (\bibinfo{year}{2017}).
\newblock


\bibitem[\protect\citeauthoryear{Joshi, Chen, Liu, Weld, Zettlemoyer, and
  Levy}{Joshi et~al\mbox{.}}{2019}]%
        {joshi2019spanbert}
\bibfield{author}{\bibinfo{person}{Mandar Joshi}, \bibinfo{person}{Danqi Chen},
  \bibinfo{person}{Yinhan Liu}, \bibinfo{person}{Daniel~S. Weld},
  \bibinfo{person}{Luke Zettlemoyer}, {and} \bibinfo{person}{Omer Levy}.}
  \bibinfo{year}{2019}\natexlab{}.
\newblock \showarticletitle{{SpanBERT}: Improving Pre-training by Representing
  and Predicting Spans}.
\newblock \bibinfo{journal}{\emph{arXiv preprint arXiv:1907.10529}}
  (\bibinfo{year}{2019}).
\newblock


\bibitem[\protect\citeauthoryear{Lan, Qiu, He, and Xu}{Lan
  et~al\mbox{.}}{2017}]%
        {lan2017continuously}
\bibfield{author}{\bibinfo{person}{Wuwei Lan}, \bibinfo{person}{Siyu Qiu},
  \bibinfo{person}{Hua He}, {and} \bibinfo{person}{Wei Xu}.}
  \bibinfo{year}{2017}\natexlab{}.
\newblock \showarticletitle{A Continuously Growing Dataset of Sentential
  Paraphrases}. In \bibinfo{booktitle}{\emph{Proceedings of The 2017 Conference
  on Empirical Methods on Natural Language Processing (EMNLP)}}.
  \bibinfo{publisher}{Association for Computational Linguistics},
  \bibinfo{pages}{1235--1245}.
\newblock


\bibitem[\protect\citeauthoryear{Lan and Xu}{Lan and Xu}{2018}]%
        {lan2018neural}
\bibfield{author}{\bibinfo{person}{Wuwei Lan} {and} \bibinfo{person}{Wei Xu}.}
  \bibinfo{year}{2018}\natexlab{}.
\newblock \showarticletitle{Neural Network Models for Paraphrase
  Identification, Semantic Textual Similarity, Natural Language Inference, and
  Question Answering}. In \bibinfo{booktitle}{\emph{Proceedings of the 27th
  International Conference on Computational Linguistics}}.
  \bibinfo{publisher}{Association for Computational Linguistics},
  \bibinfo{pages}{3890--3902}.
\newblock


\bibitem[\protect\citeauthoryear{Lan, Chen, Goodman, Gimpel, Sharma, and
  Soricut}{Lan et~al\mbox{.}}{2019}]%
        {lan2019albert}
\bibfield{author}{\bibinfo{person}{Zhenzhong Lan}, \bibinfo{person}{Mingda
  Chen}, \bibinfo{person}{Sebastian Goodman}, \bibinfo{person}{Kevin Gimpel},
  \bibinfo{person}{Piyush Sharma}, {and} \bibinfo{person}{Radu Soricut}.}
  \bibinfo{year}{2019}\natexlab{}.
\newblock \showarticletitle{ALBERT: A Lite BERT for Self-supervised Learning of
  Language Representations}. In \bibinfo{booktitle}{\emph{International
  Conference on Learning Representations}}.
\newblock


\bibitem[\protect\citeauthoryear{Li, Zhang, and Jia}{Li et~al\mbox{.}}{2018}]%
        {li2018attention}
\bibfield{author}{\bibinfo{person}{Guanyu Li}, \bibinfo{person}{Pengfei Zhang},
  {and} \bibinfo{person}{Caiyan Jia}.} \bibinfo{year}{2018}\natexlab{}.
\newblock \showarticletitle{Attention Boosted Sequential Inference Model}.
\newblock \bibinfo{journal}{\emph{CoRR}}  \bibinfo{volume}{abs/1812.01840}
  (\bibinfo{year}{2018}).
\newblock


\bibitem[\protect\citeauthoryear{Liu, Zhou, Zhao, Wang, Ju, Deng, and Wang}{Liu
  et~al\mbox{.}}{2020}]%
        {weijie2019kbert}
\bibfield{author}{\bibinfo{person}{Weijie Liu}, \bibinfo{person}{Peng Zhou},
  \bibinfo{person}{Zhe Zhao}, \bibinfo{person}{Zhiruo Wang},
  \bibinfo{person}{Qi Ju}, \bibinfo{person}{Haotang Deng}, {and}
  \bibinfo{person}{Ping Wang}.} \bibinfo{year}{2020}\natexlab{}.
\newblock \showarticletitle{K-BERT: Enabling Language Representation with
  Knowledge Graph}. In \bibinfo{booktitle}{\emph{Proceedings of AAAI 2020}}.
\newblock


\bibitem[\protect\citeauthoryear{Liu, Ott, Goyal, Du, Joshi, Chen, Levy, Lewis,
  Zettlemoyer, and Stoyanov}{Liu et~al\mbox{.}}{2019}]%
        {liu2019roberta}
\bibfield{author}{\bibinfo{person}{Yinhan Liu}, \bibinfo{person}{Myle Ott},
  \bibinfo{person}{Naman Goyal}, \bibinfo{person}{Jingfei Du},
  \bibinfo{person}{Mandar Joshi}, \bibinfo{person}{Danqi Chen},
  \bibinfo{person}{Omer Levy}, \bibinfo{person}{Mike Lewis},
  \bibinfo{person}{Luke Zettlemoyer}, {and} \bibinfo{person}{Veselin
  Stoyanov}.} \bibinfo{year}{2019}\natexlab{}.
\newblock \showarticletitle{RoBERTa: A Robustly Optimized BERT Pretraining
  Approach}.
\newblock \bibinfo{journal}{\emph{arXiv preprint arXiv:1907.11692}}
  (\bibinfo{year}{2019}).
\newblock


\bibitem[\protect\citeauthoryear{Miller}{Miller}{1995}]%
        {miller1995wordnet}
\bibfield{author}{\bibinfo{person}{George~A Miller}.}
  \bibinfo{year}{1995}\natexlab{}.
\newblock \showarticletitle{WordNet: a lexical database for English}.
\newblock \bibinfo{journal}{\emph{Commun. ACM}} \bibinfo{volume}{38},
  \bibinfo{number}{11} (\bibinfo{year}{1995}), \bibinfo{pages}{39--41}.
\newblock


\bibitem[\protect\citeauthoryear{Parikh, T{\"{a}}ckstr{\"{o}}m, Das, and
  Uszkoreit}{Parikh et~al\mbox{.}}{2016}]%
        {parikh2016decomposable}
\bibfield{author}{\bibinfo{person}{Ankur~P. Parikh}, \bibinfo{person}{Oscar
  T{\"{a}}ckstr{\"{o}}m}, \bibinfo{person}{Dipanjan Das}, {and}
  \bibinfo{person}{Jakob Uszkoreit}.} \bibinfo{year}{2016}\natexlab{}.
\newblock \showarticletitle{A Decomposable Attention Model for Natural Language
  Inference}. In \bibinfo{booktitle}{\emph{{EMNLP}}}. \bibinfo{publisher}{The
  Association for Computational Linguistics}, \bibinfo{pages}{2249--2255}.
\newblock


\bibitem[\protect\citeauthoryear{Pennington, Socher, and Manning}{Pennington
  et~al\mbox{.}}{2014}]%
        {pennington2014glove}
\bibfield{author}{\bibinfo{person}{Jeffrey Pennington},
  \bibinfo{person}{Richard Socher}, {and} \bibinfo{person}{Christopher~D
  Manning}.} \bibinfo{year}{2014}\natexlab{}.
\newblock \showarticletitle{Glove: Global vectors for word representation}. In
  \bibinfo{booktitle}{\emph{Proceedings of the 2014 conference on empirical
  methods in natural language processing (EMNLP)}}.
  \bibinfo{pages}{1532--1543}.
\newblock


\bibitem[\protect\citeauthoryear{Peters, Neumann, Iyyer, Gardner, Clark, Lee,
  and Zettlemoyer}{Peters et~al\mbox{.}}{2018}]%
        {peters2018deep}
\bibfield{author}{\bibinfo{person}{Matthew~E Peters}, \bibinfo{person}{Mark
  Neumann}, \bibinfo{person}{Mohit Iyyer}, \bibinfo{person}{Matt Gardner},
  \bibinfo{person}{Christopher Clark}, \bibinfo{person}{Kenton Lee}, {and}
  \bibinfo{person}{Luke Zettlemoyer}.} \bibinfo{year}{2018}\natexlab{}.
\newblock \showarticletitle{Deep contextualized word representations}. In
  \bibinfo{booktitle}{\emph{Proceedings of NAACL-HLT}}.
  \bibinfo{pages}{2227--2237}.
\newblock


\bibitem[\protect\citeauthoryear{Radford, Narasimhan, Salimans, and
  Sutskever}{Radford et~al\mbox{.}}{2018}]%
        {radford2018improving}
\bibfield{author}{\bibinfo{person}{Alec Radford}, \bibinfo{person}{Karthik
  Narasimhan}, \bibinfo{person}{Tim Salimans}, {and} \bibinfo{person}{Ilya
  Sutskever}.} \bibinfo{year}{2018}\natexlab{}.
\newblock \bibinfo{title}{Improving language understanding by generative
  pre-training}.
\newblock
\newblock


\bibitem[\protect\citeauthoryear{Rogers, Kovaleva, and Rumshisky}{Rogers
  et~al\mbox{.}}{2020}]%
        {rogers2020primer}
\bibfield{author}{\bibinfo{person}{Anna Rogers}, \bibinfo{person}{Olga
  Kovaleva}, {and} \bibinfo{person}{Anna Rumshisky}.}
  \bibinfo{year}{2020}\natexlab{}.
\newblock \showarticletitle{A primer in bertology: What we know about how bert
  works}.
\newblock \bibinfo{journal}{\emph{arXiv preprint arXiv:2002.12327}}
  (\bibinfo{year}{2020}).
\newblock


\bibitem[\protect\citeauthoryear{Sultan, Bethard, and Sumner}{Sultan
  et~al\mbox{.}}{2014}]%
        {sultan2014back}
\bibfield{author}{\bibinfo{person}{Md~Arafat Sultan}, \bibinfo{person}{Steven
  Bethard}, {and} \bibinfo{person}{Tamara Sumner}.}
  \bibinfo{year}{2014}\natexlab{}.
\newblock \showarticletitle{Back to Basics for Monolingual Alignment:
  Exploiting Word Similarity and Contextual Evidence}.
\newblock \bibinfo{journal}{\emph{Transactions of the Association for
  Computational Linguistics}}  \bibinfo{volume}{2} (\bibinfo{year}{2014}),
  \bibinfo{pages}{219--230}.
\newblock


\bibitem[\protect\citeauthoryear{Sultan, Bethard, and Sumner}{Sultan
  et~al\mbox{.}}{2015}]%
        {sultan2015dls}
\bibfield{author}{\bibinfo{person}{Md~Arafat Sultan}, \bibinfo{person}{Steven
  Bethard}, {and} \bibinfo{person}{Tamara Sumner}.}
  \bibinfo{year}{2015}\natexlab{}.
\newblock \showarticletitle{Dls@ cu: Sentence similarity from word alignment
  and semantic vector composition}. In \bibinfo{booktitle}{\emph{Proceedings of
  the 9th International Workshop on Semantic Evaluation (SemEval 2015)}}.
  \bibinfo{pages}{148--153}.
\newblock


\bibitem[\protect\citeauthoryear{Tenney, Das, and Pavlick}{Tenney
  et~al\mbox{.}}{2019a}]%
        {tenney2019bert}
\bibfield{author}{\bibinfo{person}{Ian Tenney}, \bibinfo{person}{Dipanjan Das},
  {and} \bibinfo{person}{Ellie Pavlick}.} \bibinfo{year}{2019}\natexlab{a}.
\newblock \showarticletitle{BERT Rediscovers the Classical NLP Pipeline}. In
  \bibinfo{booktitle}{\emph{Proceedings of the 57th Annual Meeting of the
  Association for Computational Linguistics}}. \bibinfo{pages}{4593--4601}.
\newblock


\bibitem[\protect\citeauthoryear{Tenney, Xia, Chen, Wang, Poliak, McCoy, Kim,
  Van~Durme, Bowman, Das, et~al\mbox{.}}{Tenney et~al\mbox{.}}{2019b}]%
        {tenney2019you}
\bibfield{author}{\bibinfo{person}{Ian Tenney}, \bibinfo{person}{Patrick Xia},
  \bibinfo{person}{Berlin Chen}, \bibinfo{person}{Alex Wang},
  \bibinfo{person}{Adam Poliak}, \bibinfo{person}{R~Thomas McCoy},
  \bibinfo{person}{Najoung Kim}, \bibinfo{person}{Benjamin Van~Durme},
  \bibinfo{person}{Samuel~R Bowman}, \bibinfo{person}{Dipanjan Das},
  {et~al\mbox{.}}} \bibinfo{year}{2019}\natexlab{b}.
\newblock \showarticletitle{What do you learn from context? probing for
  sentence structure in contextualized word representations}.
\newblock \bibinfo{journal}{\emph{arXiv preprint arXiv:1905.06316}}
  (\bibinfo{year}{2019}).
\newblock


\bibitem[\protect\citeauthoryear{Tomar, Duque, T{\"{a}}ckstr{\"{o}}m,
  Uszkoreit, and Das}{Tomar et~al\mbox{.}}{2017}]%
        {tomar2017neural}
\bibfield{author}{\bibinfo{person}{Gaurav~Singh Tomar}, \bibinfo{person}{Thyago
  Duque}, \bibinfo{person}{Oscar T{\"{a}}ckstr{\"{o}}m}, \bibinfo{person}{Jakob
  Uszkoreit}, {and} \bibinfo{person}{Dipanjan Das}.}
  \bibinfo{year}{2017}\natexlab{}.
\newblock \showarticletitle{Neural Paraphrase Identification of Questions with
  Noisy Pretraining}. In \bibinfo{booktitle}{\emph{SWCN@EMNLP}}.
  \bibinfo{publisher}{Association for Computational Linguistics},
  \bibinfo{pages}{142--147}.
\newblock


\bibitem[\protect\citeauthoryear{Vaswani, Shazeer, Parmar, Uszkoreit, Jones,
  Gomez, Kaiser, and Polosukhin}{Vaswani et~al\mbox{.}}{2017}]%
        {vaswani2017attention}
\bibfield{author}{\bibinfo{person}{Ashish Vaswani}, \bibinfo{person}{Noam
  Shazeer}, \bibinfo{person}{Niki Parmar}, \bibinfo{person}{Jakob Uszkoreit},
  \bibinfo{person}{Llion Jones}, \bibinfo{person}{Aidan~N Gomez},
  \bibinfo{person}{{\L}ukasz Kaiser}, {and} \bibinfo{person}{Illia
  Polosukhin}.} \bibinfo{year}{2017}\natexlab{}.
\newblock \showarticletitle{Attention is all you need}. In
  \bibinfo{booktitle}{\emph{Advances in neural information processing
  systems}}. \bibinfo{pages}{5998--6008}.
\newblock


\bibitem[\protect\citeauthoryear{Wang, Singh, Michael, Hill, Levy, and
  Bowman}{Wang et~al\mbox{.}}{2019}]%
        {wang2019glue}
\bibfield{author}{\bibinfo{person}{Alex Wang}, \bibinfo{person}{Amanpreet
  Singh}, \bibinfo{person}{Julian Michael}, \bibinfo{person}{Felix Hill},
  \bibinfo{person}{Omer Levy}, {and} \bibinfo{person}{Samuel~R. Bowman}.}
  \bibinfo{year}{2019}\natexlab{}.
\newblock \showarticletitle{{GLUE}: A Multi-Task Benchmark and Analysis
  Platform for Natural Language Understanding}.
\newblock
\newblock
\shownote{In the Proceedings of ICLR.}


\bibitem[\protect\citeauthoryear{Wang, Hamza, and Florian}{Wang
  et~al\mbox{.}}{2017}]%
        {wang2017bilateral}
\bibfield{author}{\bibinfo{person}{Zhiguo Wang}, \bibinfo{person}{Wael Hamza},
  {and} \bibinfo{person}{Radu Florian}.} \bibinfo{year}{2017}\natexlab{}.
\newblock \showarticletitle{Bilateral multi-perspective matching for natural
  language sentences}. In \bibinfo{booktitle}{\emph{Proceedings of the 26th
  International Joint Conference on Artificial Intelligence}}.
  \bibinfo{pages}{4144--4150}.
\newblock


\bibitem[\protect\citeauthoryear{Wu and Palmer}{Wu and Palmer}{1994}]%
        {wu1994verbs}
\bibfield{author}{\bibinfo{person}{Zhibiao Wu} {and} \bibinfo{person}{Martha
  Palmer}.} \bibinfo{year}{1994}\natexlab{}.
\newblock \showarticletitle{Verbs semantics and lexical selection}. In
  \bibinfo{booktitle}{\emph{Proceedings of the 32nd annual meeting on
  Association for Computational Linguistics}}. \bibinfo{pages}{133--138}.
\newblock


\bibitem[\protect\citeauthoryear{Xie, Dai, Hovy, Luong, and Le}{Xie
  et~al\mbox{.}}{2019}]%
        {xie2019unsupervised}
\bibfield{author}{\bibinfo{person}{Qizhe Xie}, \bibinfo{person}{Zihang Dai},
  \bibinfo{person}{Eduard Hovy}, \bibinfo{person}{Minh-Thang Luong}, {and}
  \bibinfo{person}{Quoc~V Le}.} \bibinfo{year}{2019}\natexlab{}.
\newblock \showarticletitle{Unsupervised data augmentation for consistency
  training}.
\newblock \bibinfo{journal}{\emph{arXiv preprint arXiv:1904.12848}}
  (\bibinfo{year}{2019}).
\newblock


\bibitem[\protect\citeauthoryear{Xu, Ritter, Callison-Burch, Dolan, and Ji}{Xu
  et~al\mbox{.}}{2014}]%
        {xu2014extracting}
\bibfield{author}{\bibinfo{person}{Wei Xu}, \bibinfo{person}{Alan Ritter},
  \bibinfo{person}{Chris Callison-Burch}, \bibinfo{person}{William~B Dolan},
  {and} \bibinfo{person}{Yangfeng Ji}.} \bibinfo{year}{2014}\natexlab{}.
\newblock \showarticletitle{Extracting lexically divergent paraphrases from
  Twitter}.
\newblock \bibinfo{journal}{\emph{Transactions of the Association for
  Computational Linguistics}}  \bibinfo{volume}{2} (\bibinfo{year}{2014}),
  \bibinfo{pages}{435--448}.
\newblock


\bibitem[\protect\citeauthoryear{Yang, Dai, Yang, Carbonell, Salakhutdinov, and
  Le}{Yang et~al\mbox{.}}{2019}]%
        {yang2019xlnet}
\bibfield{author}{\bibinfo{person}{Zhilin Yang}, \bibinfo{person}{Zihang Dai},
  \bibinfo{person}{Yiming Yang}, \bibinfo{person}{Jaime Carbonell},
  \bibinfo{person}{Russ~R Salakhutdinov}, {and} \bibinfo{person}{Quoc~V Le}.}
  \bibinfo{year}{2019}\natexlab{}.
\newblock \showarticletitle{Xlnet: Generalized autoregressive pretraining for
  language understanding}. In \bibinfo{booktitle}{\emph{Advances in neural
  information processing systems}}. \bibinfo{pages}{5753--5763}.
\newblock


\bibitem[\protect\citeauthoryear{Yin, Sch{\"u}tze, Xiang, and Zhou}{Yin
  et~al\mbox{.}}{2016}]%
        {yin2016abcnn}
\bibfield{author}{\bibinfo{person}{Wenpeng Yin}, \bibinfo{person}{Hinrich
  Sch{\"u}tze}, \bibinfo{person}{Bing Xiang}, {and} \bibinfo{person}{Bowen
  Zhou}.} \bibinfo{year}{2016}\natexlab{}.
\newblock \showarticletitle{Abcnn: Attention-based convolutional neural network
  for modeling sentence pairs}.
\newblock \bibinfo{journal}{\emph{Transactions of the Association for
  Computational Linguistics}}  \bibinfo{volume}{4} (\bibinfo{year}{2016}),
  \bibinfo{pages}{259--272}.
\newblock


\bibitem[\protect\citeauthoryear{Zhang, Han, Liu, Jiang, Sun, and Liu}{Zhang
  et~al\mbox{.}}{2019}]%
        {zhang2019ernie}
\bibfield{author}{\bibinfo{person}{Zhengyan Zhang}, \bibinfo{person}{Xu Han},
  \bibinfo{person}{Zhiyuan Liu}, \bibinfo{person}{Xin Jiang},
  \bibinfo{person}{Maosong Sun}, {and} \bibinfo{person}{Qun Liu}.}
  \bibinfo{year}{2019}\natexlab{}.
\newblock \showarticletitle{{ERNIE}: Enhanced Language Representation with
  Informative Entities}. In \bibinfo{booktitle}{\emph{Proceedings of ACL
  2019}}.
\newblock


\bibitem[\protect\citeauthoryear{Zhang, Wu, Zhao, Li, Zhang, Zhou, and
  Zhou}{Zhang et~al\mbox{.}}{2020}]%
        {zhang2019semantics}
\bibfield{author}{\bibinfo{person}{Zhuosheng Zhang}, \bibinfo{person}{Yuwei
  Wu}, \bibinfo{person}{Hai Zhao}, \bibinfo{person}{Zuchao Li},
  \bibinfo{person}{Shuailiang Zhang}, \bibinfo{person}{Xi Zhou}, {and}
  \bibinfo{person}{Xiang Zhou}.} \bibinfo{year}{2020}\natexlab{}.
\newblock \showarticletitle{Semantics-Aware {BERT} for Language Understanding}.
  In \bibinfo{booktitle}{\emph{{AAAI}}}. \bibinfo{publisher}{{AAAI} Press},
  \bibinfo{pages}{9628--9635}.
\newblock


\end{thebibliography}
\end{document}